\documentclass[runningheads]{llncs}

\usepackage{eccv}

\usepackage{eccvabbrv}

\usepackage{graphicx}
\usepackage{booktabs}
\usepackage{bm}
\usepackage{wrapfig}
\usepackage{adjustbox}
\usepackage{algorithm}  
\usepackage{algpseudocode}

\usepackage[utf8]{inputenc} 
\usepackage[T1]{fontenc}    
\usepackage{hyperref}       
\usepackage{url}            
\usepackage{booktabs}       
\usepackage{amsfonts}  

\usepackage{nicefrac}       
\usepackage{microtype}      
\usepackage{xcolor}
\usepackage{enumitem}
\usepackage{multirow}
\usepackage{mathtools}
\usepackage{amssymb}
\usepackage{colortbl}
\usepackage{graphicx}
\usepackage{url}
\usepackage{pgfplots}  
\usepackage{caption}     
\usepackage{subcaption} 
\usepackage{amsmath}
\usepackage{booktabs}
\usepackage{wrapfig}
\usepackage{float}
\usepackage{enumitem}

\usepackage[accsupp]{axessibility}  

\usepackage{hyperref}

\usepackage{orcidlink}

\begin{document}

\title{MMLoP: Multi-Modal Low-Rank Prompting for Efficient
Vision-Language Adaptation\thanks{To appear in the Proceedings of the
European Conference on Computer Vision (ECCV) 2026.}}

\titlerunning{MMLoP: Multi-Modal Low-Rank Prompting}


\author{Sajjad Ghiasvand\inst{1}\orcidlink{0009-0005-3834-7418} \and
Haniyeh Ehsani Oskouie\inst{2}\orcidlink{0009-0006-7189-3784} \and
Mahnoosh Alizadeh\inst{1}\orcidlink{0000-0003-3369-3846}\and 
Ramtin Pedarsani\inst{1}\orcidlink{0000-0003-0786-2132}}

\authorrunning{S.~Ghiasvand et al.}

\institute{University of California, Santa Barbara\\ \email{\{sajjad,alizadeh,ramtin\}@ucsb.edu}\and
University of California, Los Angeles \\\email{haniyeh@cs.ucla.edu}
}

\maketitle

\begin{abstract}
Prompt learning has become a dominant paradigm for adapting vision-language 
models (VLMs) such as CLIP to downstream tasks without modifying pretrained 
weights. While extending prompts to both vision and text encoders across multiple 
transformer layers significantly boosts performance, it dramatically increases 
the number of trainable parameters, with state-of-the-art methods requiring 
millions of parameters and abandoning the parameter efficiency that makes prompt 
tuning attractive. In this work, we propose \textbf{MMLoP} 
(\textbf{M}ulti-\textbf{M}odal \textbf{Lo}w-Rank \textbf{P}rompting), a 
framework that achieves deep multi-modal prompting with only \textbf{11.5K 
trainable parameters}, comparable to early text-only methods like CoOp. MMLoP 
parameterizes vision and text prompts at each transformer layer through a low-rank factorization that constrains prompts to a compact subspace, 
providing parameter efficiency while motivating the need for our 
complementary regularization components.
To further close the accuracy gap with 
state-of-the-art methods, we introduce three complementary components: a 
self-regulating consistency loss that anchors prompted representations to 
frozen zero-shot CLIP features at both the feature and logit levels, a uniform 
drift correction that removes the global embedding shift induced by prompt tuning 
to preserve class-discriminative structure, and a shared up-projection that couples vision and text prompts through a common low-rank factor to enforce cross-modal alignment. Extensive experiments across three benchmarks and 11 diverse datasets demonstrate that MMLoP achieves a highly favorable 
accuracy-efficiency tradeoff, outperforming the majority of existing methods including those with orders of magnitude more parameters, while achieving a harmonic mean of 79.70\% on base-to-novel generalization. Code is available at \url{https://github.com/sajjad-ucsb/MMLoP}.

\keywords{Vision-Language Models \and Prompt Learning \and Parameter Efficiency 
\and Low-Rank Adaptation \and Few-Shot Learning}
\end{abstract}

\section{Introduction}

Large-scale vision-language models (VLMs) such as CLIP~\cite{radford2021learning} 
and ALIGN~\cite{jia2021scaling} have emerged as powerful foundations for 
multi-modal understanding, enabling strong zero-shot transfer across a wide range 
of downstream tasks including image classification~\cite{zhou2022detecting}, 
object detection~\cite{li2024learning, maaz2022class,mahjourian2025multimodal}, and semantic 
segmentation~\cite{li2024omg, rao2022denseclip}. Trained on hundreds of millions 
of image-text pairs via contrastive learning, these models acquire rich, 
generalizable representations that can be readily applied to new tasks without 
any additional training. However, fully fine-tuning such models on downstream 
tasks often degrades their original generalization ability, while simple linear 
probing yields suboptimal adaptation performance~\cite{khattak2023maple}. 
This has motivated the development of parameter-efficient adaptation strategies 
that preserve pretrained representations while enabling task-specific learning.

Prompt learning has emerged as a dominant paradigm for adapting VLMs without 
modifying pretrained weights~\cite{liu2023pre, jiang2020can, shin2020autoprompt}. 
Early methods such as CoOp~\cite{COOP} and CoCoOp~\cite{cocoop} optimize 
continuous context vectors in the text branch of CLIP, achieving strong few-shot 
adaptation with as few as 2K--8K trainable parameters. Subsequent works extended 
this idea to both modalities through multi-modal deep prompting~\cite{khattak2023maple, 
royconsistency, guo2025mmrl}, where separate prompt tokens are learned at every 
transformer layer of both vision and text encoders. While this consistently boosts 
performance, it comes at a steep cost: MaPLe~\cite{khattak2023maple} requires over 
3.5M trainable parameters, and more recent methods such as CoPrompt~\cite{royconsistency} 
push this even further. In pursuing higher accuracy, these methods abandon one of 
the core promises of prompt tuning: \textit{parameter efficiency}.

\begin{figure}[t]
  \centering
\includegraphics[width=1\textwidth]{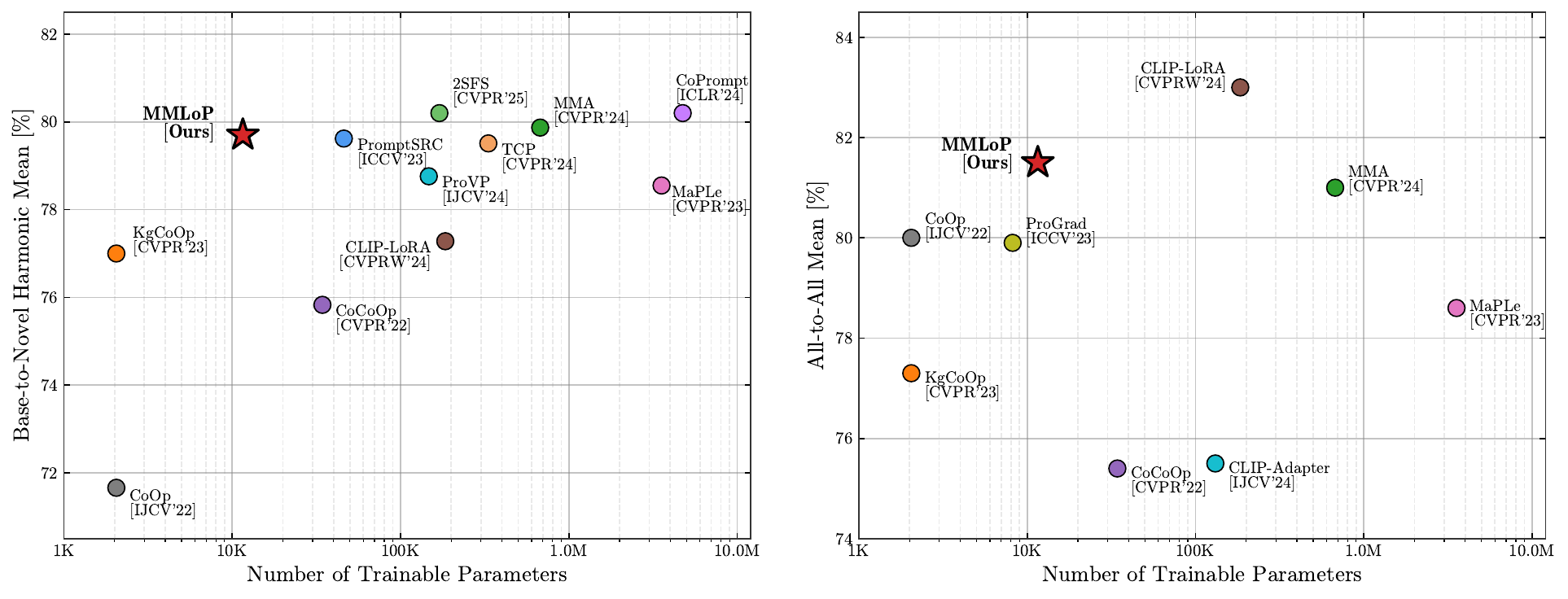}
  \caption{Accuracy vs.\ number of trainable parameters for prompt learning methods 
on base-to-novel generalization (left) and all-to-all few-shot classification 
(right).}
  \label{acc_vs_params}
\end{figure}

This tension between accuracy and efficiency motivates our work. As illustrated 
in Fig.~\ref{acc_vs_params}, existing methods that achieve competitive base-to-novel 
generalization and few-shot performance do so with orders of magnitude more 
trainable parameters than early text-only methods, while methods that remain 
parameter-efficient fall significantly short in accuracy. This raises a natural question: \textit{can deep multi-modal prompting be made as parameter-efficient as early text-only methods, without sacrificing competitive accuracy?} A natural candidate 
is low-rank factorization~\cite{hu2021lora,zanella2024low}, which has proven highly effective for 
parameter-efficient adaptation of large language models. However, naively applying 
low-rank factorization to multi-modal prompts reduces expressive capacity without 
any mechanism for cross-modal alignment, leaving a significant accuracy gap that 
must be addressed through careful regularization design.

We present \textbf{MMLoP} (\textbf{M}ulti-\textbf{M}odal \textbf{Lo}w-Rank 
\textbf{P}rompting), a parameter-efficient framework for vision-language adaptation 
that achieves competitive accuracy with only \textbf{11.5K trainable parameters} --- 
comparable to CoOp, yet benefiting from deep multi-modal prompting across both 
encoders. MMLoP parameterizes deep prompts via low-rank factorization, reducing the 
parameter count by over 300$\times$ relative to MaPLe. To compensate for the reduced 
expressiveness of the low-rank subspace, we introduce three complementary components: 
(i) a \textbf{Self-Regulating Consistency Loss} ($\mathcal{L}_{\text{SCL}}$) that 
anchors prompted features to the frozen zero-shot CLIP representations at both the 
feature and logit levels, preventing overfitting to base classes; (ii) a 
\textbf{Uniform Drift Correction} (UDC) that identifies and removes the global 
embedding shift induced by prompt tuning, preserving class-discriminative structure 
and improving generalization to novel classes; and (iii) a \textbf{Shared 
Up-Projection} that couples vision and text prompts through a common low-rank factor 
at each layer, enforcing cross-modal alignment at virtually no additional parameter cost.

Our main contributions are as follows:
\begin{itemize}
    \item We propose MMLoP, a multi-modal prompt learning framework that achieves 
    deep vision-language prompting at CoOp-level parameter cost through low-rank 
    factorization of prompt matrices across transformer layers.
    
    \item We introduce three regularization components, a self-regulating 
    consistency loss, uniform drift correction, and shared up-projection, that together recover the accuracy gap introduced by low-rank constraints while 
    improving generalization to novel classes.
    
    \item We conduct extensive experiments across three benchmarks (base-to-novel generalization, domain generalization, and all-to-all few-shot classification) on 11 diverse datasets. MMLoP outperforms 16 of 19 compared methods, including those requiring up to $\sim$300× more trainable parameters, achieving a harmonic mean of 79.70\% on base-to-novel generalization and a mean accuracy of 81.5\% on all-to-all few-shot classification, all with only 11.5K trainable parameters.
\end{itemize}

\section{Related Work}
\textbf{Vision-Language Models.} Large-scale vision-language models (VLMs) \cite{radford2021learning, jia2021scaling, yaofilip, li2022fine} have emerged as powerful foundations for multi-modal understanding by learning aligned visual and textual representations from massive image-text datasets. Models such as CLIP~\cite{radford2021learning} and ALIGN~\cite{jia2021scaling} are trained on hundreds of millions of image-text pairs using contrastive learning, enabling strong zero-shot transfer to a wide range of downstream tasks including image classification~\cite{zhou2022detecting}, object detection~\cite{li2024learning, maaz2022class}, and semantic segmentation~\cite{li2024omg}. However, fully fine-tuning these large-scale models on downstream tasks often degrades their original generalization ability, while simple linear probing yields suboptimal adaptation performance~\cite{khattak2023maple}. This has motivated the development of parameter-efficient adaptation strategies that preserve the pretrained representations while enabling task-specific learning.

\textbf{Prompt Learning.} Prompt learning, originating in NLP~\cite{shin2020autoprompt, jiang2020can, liu2023pre}, has become a dominant paradigm for adapting VLMs to downstream tasks without modifying the pretrained weights. These methods can be broadly categorized into three forms: textual prompt learning~\cite{COOP,cocoop,lu2022prompt,zhu2023prompt,liupatch,yao2023visual,yao2024tcp,park2024prompt,tian2024argue,bulat2023lasp,du2024ipo} that optimizes continuous prompt vectors in the language branch of CLIP; visual prompt learning~\cite{jia2022visual,wang2022learning,bahng2022exploring,li2024visual,yang2023fine} that introduces learnable tokens into the visual input space while keeping pretrained backbones frozen; and multi-modal prompt learning~\cite{khattak2023maple,khattak2023self,lee2023multimodal,li2023efficient,royconsistency,hao2025task,zheng2025hierarchical,yang2025learning,yao2025bi,zhang2025decouple,chengvamp,mmaghiasvand} that applies prompts to both vision and language branches for improved cross-modal alignment. Text-only methods such as CoOp~\cite{COOP}, CoCoOp~\cite{cocoop}, and KgCoOp~\cite{yao2023visual} are highly parameter-efficient but adapt only a single modality. Multi-modal approaches such as MaPLe~\cite{khattak2023maple}, CoPrompt~\cite{royconsistency}, and MMRL~\cite{guo2025mmrl} achieve stronger performance by learning deep prompts across both encoders, but at the cost of significantly increased parameter counts. \textit{In this work, we aim to bridge this gap by reducing the trainable parameter count of deep multi-modal prompting to be comparable with early text-only methods like CoOp, while maintaining competitive accuracy and generalization with state-of-the-art approaches.}

\textbf{Low-Rank Factorization for VLMs.}
Inspired by LoRA~\cite{hu2021lora}, recent work has explored low-rank decompositions for parameter-efficient adaptation of VLMs, falling into two camps depending on whether low-rank constraints are applied to backbone \emph{weight matrices}
or to learnable \emph{prompt tokens}.
Weight-space methods modify the pretrained weights during training:
CLIP-LoRA~\cite{zanella2024low} adapts the attention projections of both CLIP
encoders, Comp-LoRA~\cite{wang2025complementary} mitigates the resulting
forgetting via orthogonal gradient constraints, Block-LoRA~\cite{zhou2025one}
shares sub-matrices across layers to reduce redundancy, and AdvCLIP-LoRA~\cite{ghiasvand2025few}
extends CLIP-LoRA to the adversarial setting.
Prompt-space methods instead keep the backbone frozen:
DIP~\cite{hao2023towards} applies low-rank constraints to CLIP prompt matrices
with specialized initialization, reaching competitive accuracy with fewer than
0.5K parameters.
MMLoP also belongs to the prompt-space category, but uniquely applies low-rank
factorization to deep prompts in \emph{both} encoders and couples them via a
shared up-projection that enforces cross-modal alignment at no additional
parameter cost. \textit{Unlike CLIP-LoRA and its variants, MMLoP never modifies
the frozen CLIP backbone, retaining full zero-shot compatibility while reducing
trainable parameters by ${\sim}16\times$ relative to CLIP-LoRA and outperforming
it on base-to-novel generalization.}

\section{Preliminaries}\label{pre}
\textbf{CLIP.}
We denote the CLIP image and text encoders as $f$ and $g$, respectively, with pretrained parameters $\theta_{\text{CLIP}} = \{\theta_f, \theta_g\}$.
Given an input image $\bm{X} \in \mathbb{R}^{C \times H \times W}$, it is divided into $M$ patches and projected to produce patch tokens. A learnable class token $\bm{e}_{\text{cls}}$ is appended, forming $\tilde{\bm{X}} = \{\bm{e}_{\text{cls}}, \bm{e}_1, \bm{e}_2, \cdots, \bm{e}_M\}$. The image encoder produces a visual feature $\tilde{\bm{f}} = f(\tilde{\bm{X}}, \theta_f) \in \mathbb{R}^d$.
On the text side, a class label $c_k$ is wrapped in a template such as ``a photo of a \{class\}'', forming $\tilde{\bm{Y}}_k = \{\bm{t}_{S O S}, \bm{t}_1, \bm{t}_2, \cdots, \bm{t}_N, c_k, \bm{t}_{E O S}\}$, where $\{t_n\}_{n=1}^N$ are word embeddings of the template, and $t_{\text{SOS}}$, $t_{\text{EOS}}$ are the start and end tokens. The text encoder produces a textual feature $\tilde{\bm{g}}_k = g(\tilde{\bm{Y}}_k, \theta_g) \in \mathbb{R}^d$.
For zero-shot inference, predictions are made by matching image features with textual features of all $C$ classes:
\begin{equation}
    p(y = k \mid \bm{X}) = \frac{\exp(\text{sim}(\tilde{\bm{g}}_k, \tilde{\bm{f}}) / \tau)}{\sum_{i=1}^{C} \exp(\text{sim}(\tilde{\bm{g}}_i, \tilde{\bm{f}}) / \tau)},
\end{equation}
where $\text{sim}(\cdot, \cdot)$ denotes cosine similarity and $\tau$ is the temperature.

\textbf{Prompt Learning for CLIP.}
Instead of hand-crafted templates, prompt learning~\cite{COOP,cocoop} replaces the fixed text tokens with $T$ learnable context vectors $\bm{P}_t = \{\bm{p}_t^1, \bm{p}_t^2, \cdots, \bm{p}_t^T\}$, so that the textual input becomes
\\
$\tilde{\bm{Y}}_p = \{\bm{t}_{S O S}, \bm{P}_{\bm{t}}, \bm{t}_1, \bm{t}_2, \cdots, \bm{t}_N, c_k, \bm{t}_{E O S}\}$. The prompted textual feature is then $\tilde{\bm{g}}_p = g(\tilde{\bm{Y}}_p, \theta_g)$. Only the prompt vectors $\bm{P}_t$ are optimized via the cross-entropy loss while $\theta_{\text{CLIP}}$ remains frozen.

\textbf{Independent Vision-Language Prompting (IVLP).}
IVLP~\cite{rasheed2023fine} extends prompt learning to both modalities by appending $V$ visual prompts $\bm{P}_v = \{\bm{p}_v^1, \bm{p}_v^2, \cdots, \bm{p}_v^V\}$ and $T$ textual prompts $\bm{P}_t = \{\bm{p}_t^1, \bm{p}_t^2, \cdots, \bm{p}_t^T\}$. The image encoder processes $\tilde{\bm{X}}_p = \{\bm{P}_v, \bm{e}_{\text{cls}}, \bm{e}_1, \cdots, \bm{e}_M\}$ to produce $\tilde{\bm{f}}_p = f(\tilde{\bm{X}}_p, \theta_f)$, while the text encoder processes $\tilde{\bm{Y}}_p = \{\bm{t}_{S O S}, \bm{P}_{\bm{t}}, \bm{t}_1, \bm{t}_2, \cdots, \bm{t}_N, c_k, \bm{t}_{E O S}\}$ to produce $\tilde{\bm{g}}_p = g(\tilde{\bm{Y}}_p, \theta_g)$.
In its deep prompting variant, separate prompt sets $\bm{P}_v^{(l)}$ and $\bm{P}_t^{(l)}$ are learned at each transformer layer $l \in \{1, \cdots, L\}$. The full set of learnable parameters is $\bm{P} = \{\bm{P}_v^{(l)}, \bm{P}_t^{(l)}\}_{l=1}^L$, optimized with:
\begin{equation}
    \mathcal{L}_{\text{CE}} = \mathop{\arg\min}_{\bm{P}} \; \mathbb{E}_{(\bm{X}, y) \sim \mathcal{D}} \; \mathcal{L}\big(\text{sim}(\tilde{\bm{f}}_p, \tilde{\bm{g}}_p), \; y\big).\label{eq:CE}
\end{equation}
While effective, IVLP learns vision and text prompts \textit{independently}---the parameters $\bm{P}_v^{(l)}$ and $\bm{P}_t^{(l)}$ share no structure, providing no mechanism for cross-modal interaction during prompt optimization. This independence limits the model's ability to learn coordinated vision-language representations and can lead to overfitting on base classes at the expense of generalization to novel classes.


\section{Proposed Algorithm}
\begin{figure}[t]
  \centering
\includegraphics[width=0.75\textwidth]{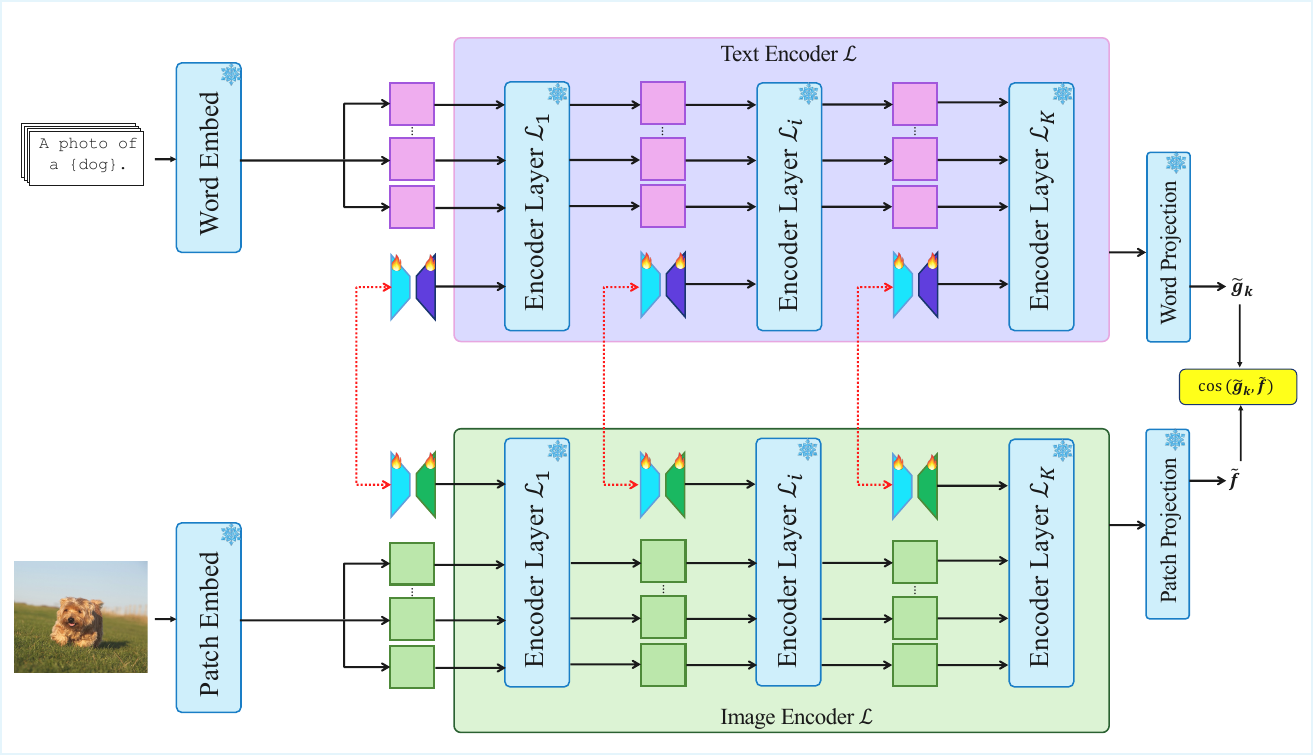}
  \caption{Overview of MMLoP. Both text and image encoders are equipped with 
deep low-rank prompts at each transformer layer, with vision 
and text prompts sharing a common up-projection matrix $\bm{U}^{(l)}$ for 
cross-modal alignment. Snowflake icons indicate frozen CLIP parameters. 
The self-regulating consistency loss is omitted 
for clarity.}
  \label{main_fig}
\end{figure}

\begin{algorithm}[t]
\caption{MMLoP: Multi-Modal Low-Rank Prompting}
\label{alg:mmlop}
\begin{algorithmic}[1]
\Require Training data $\mathcal{D}$, frozen CLIP encoders $f$, $g$, frozen zero-shot text features $\{\tilde{g}_k\}_{k=1}^C$, number of classes $C$, rank $r$, prompt depth $L$, loss weights $\lambda_1$, $\lambda_2$
\Ensure Trained low-rank factors $\{U^{(l)}, V_v^{(l)}, V_t^{(l)}\}_{l=1}^L$

\vspace{2pt}
\State \textbf{Initialize} $U^{(l)} \in \mathbb{R}^{T \times r}$, $V_v^{(l)} \in \mathbb{R}^{r \times d_v}$, $V_t^{(l)} \in \mathbb{R}^{r \times d_t}$ with $\mathcal{N}(0, 0.05)$ for $l = 1, \dots, L$

\vspace{2pt}
\For{each training iteration}
    \State Sample mini-batch $\{(x_i, y_i)\}$ from $\mathcal{D}$

    \vspace{2pt}
    \State \textcolor{gray}{\textit{// Construct low-rank prompts via shared up-projection}}
    \For{$l = 1, \dots, L$}
        \State $P_v^{(l)} \leftarrow U^{(l)} V_v^{(l)}$ \Comment{Vision prompt at layer $l$}
        \State $P_t^{(l)} \leftarrow U^{(l)} V_t^{(l)}$ \Comment{Text prompt at layer $l$}
    \EndFor

    \vspace{2pt}
    \State \textcolor{gray}{\textit{// Extract prompted features}}
    \State $\tilde{f}_p \leftarrow f\big(\{P_v^{(l)}\}, x_i\big)$, \quad $\tilde{g}_k^p \leftarrow g\big(\{P_t^{(l)}\}, c_k\big)$ for all $k$

    \vspace{2pt}
    \State \textcolor{gray}{\textit{// Uniform Drift Correction (UDC)}}
    \State $r_k \leftarrow \tilde{g}_k^p - \tilde{g}_k$ \quad for all $k$
    \State $\bar{r} \leftarrow \frac{1}{C} \sum_{k=1}^C r_k$
    \State $\hat{g}_k \leftarrow \tilde{g}_k + (r_k - \bar{r})$, \quad $\hat{g}_k \leftarrow \hat{g}_k / \|\hat{g}_k\|$ \quad for all $k$

    \vspace{2pt}
    \State \textcolor{gray}{\textit{// Compute losses}}
    \State $\mathcal{L}_{\text{CE}} \leftarrow \text{CrossEntropy}\big(\{\operatorname{sim}(\tilde{f}_p, \hat{g}_k)\}, y_i\big)$
    \State $\mathcal{L}_{\text{SCL-text}} \leftarrow \lambda_1 \|\hat{g}_k - \tilde{g}_k\|_1$
    \State $\mathcal{L}_{\text{SCL-image}} \leftarrow \lambda_2 \|\tilde{f}_p - \tilde{f}\|_1$
    \State $\mathcal{L}_{\text{SCL-logits}} \leftarrow \frac{1}{2}\mathcal{D}_{\mathcal{KL}}\big(\operatorname{sim}(\tilde{f}_p, \hat{g}_k) \| \operatorname{sim}(\tilde{f}, \tilde{g}_k)\big) + \frac{1}{2}\mathcal{D}_{\mathcal{KL}}\big(\operatorname{sim}(\tilde{f}, \tilde{g}_k) \| \operatorname{sim}(\tilde{f}_p, \hat{g}_k)\big)$
    \State $\mathcal{L} \leftarrow \mathcal{L}_{\text{CE}} + \mathcal{L}_{\text{SCL-text}} + \mathcal{L}_{\text{SCL-image}} + \mathcal{L}_{\text{SCL-logits}}$

    \vspace{2pt}
    \State Update $\{U^{(l)}, V_v^{(l)}, V_t^{(l)}\}_{l=1}^L$ via SGD on $\mathcal{L}$
\EndFor

\vspace{4pt}
\State \textbf{Inference:} compute $\hat{g}_k$ via UDC, predict $\hat{y} = \arg\max_k \operatorname{sim}(\tilde{f}_p, \hat{g}_k)$
\end{algorithmic}
\end{algorithm}

\subsection{Motivation}
Deep multi-modal prompting~\cite{khattak2023maple, yang2024mma, guo2025mmrl} 
significantly boosts few-shot recognition performance over text-only 
methods~\cite{COOP, yao2023visual}, but dramatically increases the number of 
trainable parameters --- MaPLe~\cite{khattak2023maple} requires over 3.5M 
parameters, while CoPrompt~\cite{royconsistency} pushes this even further. In pursuing 
higher accuracy, these approaches abandon one of the core promises of prompt 
tuning: \emph{parameter efficiency}. MMLoP addresses this by parameterizing deep 
prompts through a low-rank factorization inspired by~\cite{hu2021lora,ghiasvand2025few}, reducing 
the parameter count to CoOp-level while maintaining competitive performance. To 
further close the gap with state-of-the-art methods, we introduce three key 
components: (i) a \emph{self-regulating consistency loss} that prevents the 
prompted model from drifting away from CLIP's pretrained representations, (ii) a 
\emph{uniform drift correction} that removes the global embedding shift induced 
by prompt tuning, and (iii) a \emph{shared up-projection} that couples vision and 
text prompts through a common low-rank factor, enforcing cross-modal alignment.

\subsection{Low-Rank Prompt Parameterization}

Instead of learning full-rank prompt matrices $\bm{P}_v^{(l)} \in \mathbb{R}^{V \times d_v}$ and $\bm{P}_t^{(l)} \in \mathbb{R}^{T \times d_t}$ at each layer $l$, we parameterize the prompts through a low-rank factorization. Drawing inspiration from~\cite{hu2021lora}, we decompose each prompt matrix into the product of two low-rank factors:
\begin{align}
    \bm{P}_v^{(l)} = \bm{U}_v^{(l)} \bm{V}_v^{(l)},\quad
    \bm{P}_t^{(l)} = \bm{U}_t^{(l)} \bm{V}_t^{(l)}, \label{eq:lora}
\end{align}
where $\bm{U}_v^{(l)} \in \mathbb{R}^{V \times r}$ and $\bm{U}_t^{(l)} \in \mathbb{R}^{T \times r}$ are the up-projection matrices, $\bm{V}_v^{(l)} \in \mathbb{R}^{r \times d_v}$ and $\bm{V}_t^{(l)} \in \mathbb{R}^{r \times d_t}$ are the down-projection matrices, and $r \ll \min(d_v, d_t)$ is the rank.

This factorization constrains each prompt to lie in a rank-$r$ subspace, 
reducing expressive capacity while providing a foundation for parameter 
efficiency. As shown in Table~\ref{tab:ablations_on_components}, this constraint alone is insufficient 
for competitive performance, which motivates our three complementary 
regularization components. The prompted image and text inputs at layer $l$ follow the same structure as IVLP:
\begin{gather}
    \tilde{\bm{X}}_p^{(l)} = \{\bm{P}_v^{(l)},\, \bm{e}_{\text{cls}}^{(l)},\, \bm{e}_1^{(l)},\, \ldots,\, \bm{e}_M^{(l)}\}, \nonumber\\
    \tilde{\bm{Y}}_p^{(l)} = \{t_{\text{SOS}}^{(l)},\, \bm{P}_t^{(l)},\, t_1^{(l)},\, \ldots,\, t_N^{(l)},\, c_k^{(l)},\, t_{\text{EOS}}^{(l)}\},\label{lora_prompts}
\end{gather}
where the superscript $(l)$ denotes representations at layer $l$ of the transformer. During training, the entire CLIP backbone $\theta_{\text{CLIP}}$ remains frozen and only the low-rank factors $\{\bm{U}_v^{(l)}, \bm{V}_v^{(l)}, \bm{U}_t^{(l)}, \bm{V}_t^{(l)}\}_{l=1}^{L}$ are optimized.\footnote{This is the general (independent up-projection) parameterization of
Eq.~\eqref{eq:lora}. In our default model, the vision and text prompts share a single
up-projection (Sec.~\ref{shared-projection}, Eq.~\eqref{eq:shared}), so the optimized set reduces to
$\{\bm{U}^{(l)}, \bm{V}_v^{(l)}, \bm{V}_t^{(l)}\}_{l=1}^{L}$, consistent with Algorithm~1.}

\subsection{Self-Regulating Consistency Loss (SCL)}

While the low-rank parameterization reduces the risk of overfitting, prompt-tuned models can still drift away from the pretrained CLIP representations, harming generalization to unseen classes. To mitigate this, we incorporate a consistency regularization inspired by~\cite{khattak2023self} that anchors the learned features to the frozen zero-shot CLIP features. Let $\tilde{\bm{f}}_p$ and $\tilde{\bm{g}}_p$ denote the image and text features produced by the prompted model, and let $\tilde{\bm{f}}$ and $\tilde{\bm{g}}$ denote the features from the frozen zero-shot CLIP model. We define three consistency terms.

\textbf{Feature-level consistency.} We penalize the deviation between the prompted and zero-shot features in both modalities using the L1 norm:
\begin{equation}
    \mathcal{L}_{\text{SCL-image}} = \| \tilde{\bm{f}}_p - \tilde{\bm{f}} \|_1, \quad
    \mathcal{L}_{\text{SCL-text}} = \| \tilde{\bm{g}}_p - \tilde{\bm{g}} \|_1.
    \label{eq:scl_features}
\end{equation}

\textbf{Logit-level consistency.} We also regularize the output logit distributions to remain close to those of the zero-shot model. Unlike \cite{khattak2023self} which employs a standard (asymmetric) KL divergence for this purpose, we adopt a symmetric KL divergence, as we find it more effective in practice:
\begin{equation}
    \mathcal{L}_{\text{SCL-logits}} = \frac{1}{2} \mathcal{D}_{\mathcal{KL}} \Big( \operatorname{sim}(\tilde{\bm{f}}_p, \tilde{\bm{g}}_p) \;\|\; \operatorname{sim}(\tilde{\bm{f}}, \tilde{\bm{g}}) \Big) + \frac{1}{2} \mathcal{D}_{\mathcal{KL}} \Big( \operatorname{sim}(\tilde{\bm{f}}, \tilde{\bm{g}}) \;\|\; \operatorname{sim}(\tilde{\bm{f}}_p, \tilde{\bm{g}}_p) \Big).
    \label{eq:scl_logits}
\end{equation}
This symmetric formulation penalizes divergence equally in both directions, treating the prompted and zero-shot distributions more uniformly and avoiding the asymmetric gradient behavior of the standard KL. An empirical comparison of both variants is provided in Table~A3 in the Appendix. The total consistency loss is:
\begin{equation}
    \mathcal{L}_{\text{SCL}} = \lambda_1 \, \mathcal{L}_{\text{SCL-text}} + \lambda_2 \, \mathcal{L}_{\text{SCL-image}} + \mathcal{L}_{\text{SCL-logits}},
    \label{eq:scl_total}
\end{equation}
where $\lambda_1$ and $\lambda_2$ are weighting hyperparameters. The overall training objective is then:
$
    \mathcal{L} = \mathcal{L}_{\text{CE}} + \mathcal{L}_{\text{SCL}}.
    \label{eq:total_loss}
$


\subsection{Uniform Drift Correction (UDC)}

While the self-regulating consistency loss encourages the prompted model to stay close to CLIP's pretrained representations, prompt tuning can still induce a systematic shift that affects all class embeddings uniformly. Such a shift does not improve discrimination between classes---it is shared across the entire embedding space and therefore carries no class-specific signal. Rather, it reflects a form of base-class bias absorbed into the prompt during few-shot training, which disproportionately harms generalization to novel classes.

To address this, we propose a correction that removes the uniform component of the learned text feature shift throughout both training and evaluation. Let $\tilde{g}_k^p$ denote the prompted text feature for class $k$, and let $\tilde{g}_k$ denote the corresponding frozen zero-shot feature. We decompose the prompted feature into a zero-shot anchor and a residual:
$
    r_k = \tilde{g}_k^p - \tilde{g}_k.
$
The mean residual $\bar{r} = \frac{1}{C}\sum_{k=1}^C r_k$ captures the uniform drift shared across all $C$ training classes. We subtract this common component and reconstruct the corrected feature:
\begin{equation}
    \hat{g}_k = \tilde{g}_k + (r_k - \bar{r}), \qquad \hat{g}_k \leftarrow \frac{\hat{g}_k}{\|\hat{g}_k\|}.
\end{equation}
The corrected features $\hat{g}_k$ retain the class-specific adaptations learned by the prompt while eliminating the shared bias. Furthermore, since UDC is applied during training, $\mathcal{L}_{\text{SCL-text}}$ is computed on drift-corrected features $\hat{g}_k$ rather than raw prompted features. This means the consistency regularization acts on the class-discriminative residuals alone, making UDC and $\mathcal{L}_{\text{SCL}}$ mutually reinforcing rather than redundant: the consistency loss encourages meaningful class-specific adaptation while UDC ensures that any global offset is continuously removed. The correction requires no additional parameters and introduces no asymptotic
overhead: it reuses the frozen zero-shot features already computed for
$\mathcal{L}_{\text{SCL}}$ and the per-iteration text-encoder forward over all
$C$ classes that every CLIP-style few-shot method already performs to compute
the cross-entropy loss, adding only an $\mathcal{O}(Cd)$ mean subtraction on top.

\subsection{Cross-Modal Coupling via Shared Up-Projection}
\label{shared-projection}

The low-rank parameterization in Eq.~\eqref{eq:lora} with independent up-projections $\bm{U}_v^{(l)}$ and $\bm{U}_t^{(l)}$ already provides parameter efficiency. However, the vision and text prompts remain structurally decoupled---each modality's low-rank subspace is learned independently. We observe that by tying the up-projection matrices across modalities, we can introduce cross-modal interaction at virtually no additional cost. Specifically, we set $T = V$ and constrain the two modalities to share a single up-projection:
\begin{align}
    \bm{P}_v^{(l)} = \bm{U}^{(l)} \bm{V}_v^{(l)},\quad
    \bm{P}_t^{(l)} = \bm{U}^{(l)} \bm{V}_t^{(l)}, \label{eq:shared}
\end{align}
where $\bm{U}^{(l)} \in \mathbb{R}^{T \times r}$ is now shared between modalities. This design means that the vision and text prompts at each layer are constrained to share the same row space. The shared factor $\bm{U}^{(l)}$ determines the token-wise activation pattern common to both modalities, while the modality-specific factors $\bm{V}_v^{(l)}$ and $\bm{V}_t^{(l)}$ project this shared structure into the respective embedding spaces.

This coupling acts as an additional regularizer: gradient updates to $\bm{U}^{(l)}$ must simultaneously benefit both modalities, which discourages overfitting to modality-specific noise in the few-shot training data. As shown in our ablation study (Table~\ref{tab:ablations_on_components}), introducing the shared up-projection on top of the consistency loss and UDC further improves novel class accuracy by $+0.86\%$ and the harmonic mean by $+0.50\%$, demonstrating that cross-modal structural alignment provides complementary benefits to feature-level regularization.


\section{Empirical Results}

\subsection{Experiments Setup}
We evaluate MMLoP on three benchmark settings: base-to-novel generalization, domain generalization, and all-to-all few-shot classification.

\textbf{Base-to-Novel Generalization.} Following the protocol established by CoOp~\cite{COOP} and CoCoOp~\cite{cocoop}, each dataset is equally split into base and novel classes. The model is trained exclusively on base classes with 16 shots per class and evaluated on both base and novel classes. The harmonic mean (HM) of base and novel accuracy is reported as the primary metric. This setting evaluates whether the model can learn task-specific representations without sacrificing CLIP's inherent zero-shot generalization ability. We conduct experiments on 11 diverse image recognition datasets: ImageNet~\cite{imagenet} and Caltech101~\cite{caltech101} for generic object recognition; OxfordPets~\cite{oxford_pets}, StanfordCars~\cite{stanford_cars}, Flowers102~\cite{flowers102}, Food101~\cite{food101}, and FGVCAircraft~\cite{maji2013fine} for fine-grained classification; SUN397~\cite{sun397} for scene recognition; UCF101~\cite{ucf101} for action recognition; DTD~\cite{dtd} for texture classification; and EuroSAT~\cite{eurosat} for satellite image classification.

\begin{table}[tp]
\centering
\setlength{\abovecaptionskip}{0.15cm}  
\caption{Base-to-novel generalization results across multiple datasets. HM refers to harmonic mean.}
\label{tab:base_to_novel}
\renewcommand\arraystretch{1.1}
\setlength{\tabcolsep}{1.5mm}{
\resizebox{0.8\textwidth}{!}{
    \begin{tabular}{|l|ccc|ccc|ccc|ccc|}
    \hline
    \multirow{2}{*}{\textbf{Method}} &
      \multicolumn{3}{c|}{\textbf{Average}} &
      \multicolumn{3}{c|}{\textbf{ImageNet}} &
      \multicolumn{3}{c|}{\textbf{Caltech101}} &
      \multicolumn{3}{c|}{\textbf{OxfordPets}} \\ \cline{2-13}
     & Base & Novel & HM & Base & Novel & HM & Base & Novel & HM & Base & Novel & HM \\ \hline
    $\text{CLIP}$~\cite{radford2021learning} & 69.34 & 74.22 & 71.70 & 72.43 & 68.14 & 70.22 & 96.84 & 94.00 & 95.40 & 91.17 & 97.26 & 94.12 \\
    $\text{CoOp}$~\cite{COOP} & 82.69 & 63.22 & 71.66 & 76.47 & 67.88 & 71.92 & 98.00 & 89.81 & 93.73 & 93.67 & 95.29 & 94.47 \\
    $\text{CoCoOp}$~\cite{cocoop} & 80.47 & 71.69 & 75.83 & 75.98 & 70.43 & 73.10 & 97.96 & 93.81 & 95.84 & 95.20 & 97.69 & 96.43 \\
    $\text{ProDA}$~\cite{lu2022prompt} & 81.56 & 72.30 & 76.65 & 75.40 & 70.23 & 72.72 & 98.27 & 93.23 & 95.68 & 95.43 & 97.83 & 96.62 \\
    $\text{ProGrad}$~\cite{zhu2023prompt} & 82.48 & 70.75 & 76.16 & 77.02 & 66.66 & 71.46 & 98.02 & 93.89 & 95.91 & 95.07 & 97.63 & 96.33 \\
    $\text{KgCoOp}$~\cite{yao2023visual} & 80.73 & 73.60 & 77.00 & 75.83 & 69.96 & 72.78 & 97.72 & 94.39 & 96.03 & 94.65 & 97.76 & 96.18 \\
    $\text{MaPLe}$~\cite{khattak2023maple} & 82.28 & 75.14 & 78.55 & 76.66 & 70.54 & 73.47 & 97.74 & 94.36 & 96.02 & 95.43 & 97.76 & 96.58 \\
    $\text{IVLP}$~\cite{rasheed2023fine} & 84.21 & 71.79  & 77.51 & 77.00 & 66.50 & 71.37 & 98.30 & 93.20 & 95.68 & 94.90 & 97.20 & 96.04 \\
    $\text{PromptSRC}$~\cite{khattak2023self} & 84.23 &  75.48 & 79.62 & 77.60  & 70.73 & 74.01 & 97.93 & 93.83 &  95.84 &  95.47 & 97.40 & 96.43 \\
    $\text{LASP}$~\cite{bulat2023lasp} & 82.70 & 74.90 & 78.61 & 76.20 & 70.95 & 73.48 & 98.10 & 94.24 & 96.16 & 95.90 & 97.93 & 96.90 \\
    $\text{RPO}$~\cite{lee2023read} & 81.13 & 75.00 & 77.78 & 76.60 & 71.57 & 74.00 & 97.97 & 94.37 & 96.03 & 94.63 & 97.50 & 96.05 \\
    $\text{ProVP}$~\cite{xu2025progressive} & 85.20 & 73.22 & 78.76 & 75.82 & 69.21 & 72.36 & 98.92 & 94.21 & 96.51 & 95.87 & 97.65 & 96.75 \\
    $\text{LoGoPrompt}$~\cite{shi2023logoprompt} & 84.47 & 74.24 & 79.03 & 76.74 & 70.83 & 73.66 & 98.19 & 93.78 & 95.93 & 96.07 & 96.31 & 96.18 \\
    $\text{MetaPrompt}$~\cite{zhao2024learning} & 83.65 & 75.48 & 79.09 & 77.52 & 70.83 & 74.02 & 98.13 & 94.58 & 96.32 & 95.53 & 97.00 & 96.26 \\
    $\text{CLIP-LoRA}$~\cite{zanella2024low} & 85.32 & 70.63 & 77.28 & 77.58 & 68.76 & 72.91 & 98.19 & 93.05 & 95.55 & 94.36 & 95.71 & 95.03 \\
    $\text{TCP}$~\cite{yao2024tcp} & 84.13 & 75.36 & 79.51 & 77.27 & 69.87 & 73.38 & 98.23 & {94.67} & 96.42 & 94.67 & 97.20 & 95.92 \\
    $\text{MMA}$~\cite{yang2024mma} & 83.20 & 76.80 & 79.87 & 77.31 & 71.00 & 74.02 & 98.40 & 94.00 & 96.15 & 95.40 & 98.07 & 96.72 \\
    $\text{CoPrompt}$~\cite{royconsistency} & 84.00 & 77.23 & 80.48 & 77.67 & 71.27 & 74.33 & 98.27 & 94.90 & 96.55 & 95.67 & 98.10 & 96.87 \\
    $\text{2SFS}$~\cite{farina2025rethinking} & 85.55 & 75.48 & 80.20 & 77.71 & 70.99 & 74.20 & 98.71 & 94.43 & 96.52 & 95.32 & 97.82 & 96.55 \\
    \cline{1-13}
    \rowcolor{lightgray!30}
    MMLoP (Ours) & 83.79 & 75.98 & 79.70 & 77.00 & 70.50 & 73.61 & 98.27 & 93.93 & 96.05 & 95.47 & 96.97 & 96.21 \\ \hline
    \hline
    \multirow{2}{*}{\textbf{Method}} &
      \multicolumn{3}{c|}{\textbf{StanfordCars}} &
      \multicolumn{3}{c|}{\textbf{Flowers102}} &
      \multicolumn{3}{c|}{\textbf{Food101}} &
      \multicolumn{3}{c|}{\textbf{FGVCAircraft}} \\ \cline{2-13}
     & Base & Novel & HM & Base & Novel & HM & Base & Novel & HM & Base & Novel & HM \\ \hline
    $\text{CLIP}$~\cite{radford2021learning} & 63.37 & 74.89 & 68.65 & 72.08 & 77.80 & 74.83 & 90.10 & 91.22 & 90.66 & 27.19 & 36.29 & 31.09 \\
    $\text{CoOp}$~\cite{COOP} & 78.12 & 60.40 & 68.13 & 97.60 & 59.67 & 74.06 & 88.33 & 82.26 & 85.19 & 40.44 & 22.30 & 28.75 \\
    $\text{CoCoOp}$~\cite{cocoop} & 70.49 & 73.59 & 72.01 & 94.87 & 71.75 & 81.71 & 90.70 & 91.29 & 90.99 & 33.41 & 23.71 & 27.74 \\
    $\text{ProDA}$~\cite{lu2022prompt} & 74.70 & 71.20 & 72.91 & 97.70 & 68.68 & 80.66 & 90.30 & 88.57 & 89.43 & 36.90 & 34.13 & 35.46 \\
    $\text{ProGrad}$~\cite{zhu2023prompt} & 77.68 & 68.63 & 72.88 & 95.54 & 71.87 & 82.03 & 90.37 & 89.59 & 89.98 & 40.54 & 27.57 & 32.82 \\
    $\text{KgCoOp}$~\cite{yao2023visual} & 71.76 & 75.04 & 73.36 & 95.00 & 74.73 & 83.65 & 90.50 & 91.70 & 91.09 & 36.21 & 33.55 & 34.83 \\
    $\text{MaPLe}$~\cite{khattak2023maple} & 72.94 & 74.00 & 73.47 & 95.92 & 72.46 & 82.56 & 90.71 & 92.05 & 91.38 & 37.44 & 35.61 & 36.50 \\
    $\text{IVLP}$~\cite{rasheed2023fine} & 79.53 & 71.47 & 75.28 & 97.97 & 72.10 & 83.07 & 89.37 & 90.30 & 89.83 & 42.60 & 25.23 & 31.69 \\
    $\text{PromptSRC}$~\cite{khattak2023self} & 78.50 & 75.40  & 76.92 &  97.90 & 76.37 & 85.80 & 90.87 & 91.53 & 91.20 & 42.70 &  37.07 & 39.69 \\
    $\text{LASP}$~\cite{bulat2023lasp} & 75.17 & 71.60 & 73.34 & 97.00 & 74.00 & 83.95 & 91.20 & 91.70 & 91.44 & 34.53 & 30.57 & 32.43 \\
    $\text{RPO}$~\cite{lee2023read} & 73.87 & 75.53 & 74.69 & 94.13 & 76.67 & 84.50 & 90.33 & 90.83 & 90.58 & 37.33 & 34.20 & 35.70 \\
    $\text{ProVP}$~\cite{xu2025progressive} & 80.43 & 67.96 & 73.67 & 98.42 & 72.06 & 83.20 & 90.32 & 90.91 & 90.61 & 47.08 & 29.87 & 36.55 \\
    $\text{LoGoPrompt}$~\cite{shi2023logoprompt} & 78.36 & 72.39  & 75.26 & 99.05 & 76.52  & 86.34 & 90.82 & 91.41 & 91.11 & 45.98 & 34.67 & 39.53 \\
    $\text{MetaPrompt}$~\cite{zhao2024learning} & 76.34 & 75.01 & 75.48 & 97.66 & 74.49 & 84.52 & 90.74 & 91.85 & 91.29 & 40.14 & 36.51 & 38.24 \\
    $\text{CLIP-LoRA}$~\cite{zanella2024low} & 83.93 & 65.54 & 73.60 & 97.91 & 68.61 & 80.68 & 86.84 & 86.67 & 86.76 & 50.10 & 26.03 & 34.26 \\
    $\text{TCP}$~\cite{yao2024tcp} & 80.80 & 74.13 & 77.32 & 97.73 & 75.57 & 85.23 & 90.57 & 91.37 & 90.97 & 41.97 & 34.43 & 37.83 \\
    $\text{MMA}$~\cite{yang2024mma} & 78.50 & 73.10 & 75.70 & 97.77 & 75.93 & 85.48 & 90.13 & 91.30 & 90.71 & 40.57 & 36.33 & 38.33 \\
    $\text{CoPrompt}$~\cite{royconsistency} & 76.97 & 74.40 & 75.66 & 97.27 & 76.60 & 85.71 & 90.73 & 92.07 & 91.4 & 40.20 & 39.33 & 39.76 \\
    $\text{2SFS}$~\cite{farina2025rethinking} & 82.50 & 74.80 & 78.46 & 98.29 & 76.17 & 85.83 & 89.11 & 91.34 & 90.21 & 47.48 & 35.51 & 40.63 \\
    \cline{1-13}
    \rowcolor{lightgray!30}
    MMLoP (Ours) & 77.77 & 74.83 & 76.27 & 97.63 & 76.73 & 85.93 & 90.70 & 91.70 & 91.19 & 42.17 & 34.60 & 38.01 \\ \hline
    \hline
    \multirow{2}{*}{\textbf{Method}} &
      \multicolumn{3}{c|}{\textbf{SUN397}} &
      \multicolumn{3}{c|}{\textbf{DTD}} &
      \multicolumn{3}{c|}{\textbf{EuroSAT}} &
      \multicolumn{3}{c|}{\textbf{UCF101}} \\ \cline{2-13}
     & Base & Novel & HM & Base & Novel & HM & Base & Novel & HM & Base & Novel & HM \\ \hline
    $\text{CLIP}$~\cite{radford2021learning} & 69.36 & 75.35 & 72.23 & 53.24 & 59.90 & 56.37 & 56.48 & 64.05 & 60.03 & 70.53 & 77.50 & 73.85 \\
    $\text{CoOp}$~\cite{COOP} & 80.60 & 65.89 & 72.51 & 79.44 & 41.18 & 54.24 & 92.19 & 54.74 & 68.69 & 84.69 & 56.05 & 67.46 \\
    $\text{CoCoOp}$~\cite{cocoop} & 79.74 & 76.86 & 78.27 & 77.01 & 56.00 & 64.85 & 87.49 & 60.04 & 71.21 & 82.33 & 73.45 & 77.64 \\
    $\text{ProDA}$~\cite{lu2022prompt} & 78.67 & 76.93 & 77.79 & 80.67 & 56.48 & 66.44 & 83.90 & 66.00 & 73.88 & 85.23 & 71.97 & 78.04 \\
    $\text{ProGrad}$~\cite{zhu2023prompt} & 81.26 & 74.17 & 77.55 & 77.35 & 52.35 & 62.45 & 90.11 & 60.89 & 72.67 & 84.33 & 74.94 & 79.35 \\
    $\text{KgCoOp}$~\cite{yao2023visual} & 80.29 & 76.53 & 78.36 & 77.55 & 54.99 & 64.35 & 85.64 & 64.34 & 73.48 & 82.89 & 76.67 & 79.65 \\
    $\text{MaPLe}$~\cite{khattak2023maple} & 80.82 & 78.70 & 79.75 & 80.36 & 59.18 & 68.16 & 94.07 & 73.23 & 82.35 & 83.00 & 78.66 & 80.77 \\
    $\text{IVLP}$~\cite{rasheed2023fine} & 81.60 & 75.50 & 78.43 & 82.40 & 56.20 & 66.82 & 96.73 & 67.83 & 79.74 & 85.93 & 74.17 & 79.62 \\
    $\text{PromptSRC}$~\cite{khattak2023self} & 82.67 & 78.47  & 80.52 & 83.37 & 62.47 & 71.42 & 93.17 & 68.33 & 78.84 & 86.37 & 78.70 & 82.36 \\
    $\text{LASP}$~\cite{bulat2023lasp} & 80.70 & 78.60 & 79.63 & 81.40 & 58.60 & 68.14 & 94.60 & 77.78 & 85.36 & 84.77 & 78.03 & 81.26 \\
    $\text{RPO}$~\cite{lee2023read} & 80.60 & 77.80 & 79.18 & 76.70 & 62.13 & 68.61 & 86.63 & 68.97 & 76.79 & 83.67 & 75.43 & 79.34 \\
    $\text{ProVP}$~\cite{xu2025progressive} & 80.67 & 76.11 & 78.32 & 83.95 & 59.06 & 69.34 & 97.12 & 72.91 & 83.29 & 88.56 & 75.55 & 81.54 \\
    $\text{LoGoPrompt}$~\cite{shi2023logoprompt} & 81.20 & 78.12 & 79.63 & 82.87 & 60.14 &  69.70 & 93.67  & 69.44  & 79.75 & 86.19 & 73.07 & 79.09 \\
    $\text{MetaPrompt}$~\cite{zhao2024learning} & 82.26 & 79.04 & 80.62 & 83.10 & 58.05 & 68.35 & 93.53 & 75.21 & 83.38 & 85.33 & 77.72 & 81.35 \\
    $\text{CLIP-LoRA}$~\cite{zanella2024low} & 81.11 & 74.53 & 77.68 & 83.95 & 62.84 & 71.39 & 97.04 & 62.50 & 76.03 & 87.52 & 72.74 & 79.45 \\
    $\text{TCP}$~\cite{yao2024tcp} & 82.63 & 78.20 & 80.35 & 82.77 & 58.07 & 68.25 & 91.63 & 74.73 & 82.32 & 87.13 & {80.77} & 83.83 \\
    $\text{MMA}$~\cite{yang2024mma} & 82.27 & 78.57 & 80.38 & 83.20 & 65.63 & 73.38 & 85.46 & 82.34 & 83.87 & 86.23 & 80.03 & 82.20 \\
    $\text{CoPrompt}$~\cite{royconsistency} & 82.63 & 80.03 & 81.31 & 83.13 & 64.73 & 72.79 & 94.60 & 78.57 & 85.84 & 86.90 & 79.57 & 83.07 \\
    $\text{2SFS}$~\cite{farina2025rethinking} & 82.59 & 78.91 & 80.70 & 84.60 & 65.01 & 73.52 & 96.91 & 67.09 & 79.29 & 87.85 & 78.19 & 82.74 \\
    \cline{1-13}
    \rowcolor{lightgray!30}
    MMLoP (Ours) & 82.40 & 78.13 & 80.21 & 82.67 & 60.87 & 70.11 & 92.37 & 79.10 & 85.22 & 85.23 & 78.37 & 81.66 \\ \hline
    \end{tabular}
}
}
\end{table}

\textbf{Domain Generalization.} To assess robustness to distribution shifts, we train the model on ImageNet~\cite{imagenet} and directly evaluate on four out-of-distribution variants: ImageNetV2~\cite{recht2019imagenet}, ImageNet-Sketch~\cite{wang2019learning}, ImageNet-A~\cite{hendrycks2021natural}, and ImageNet-R~\cite{hendrycks2021many}, each introducing a different type of domain shift.

\textbf{All-to-All Few-Shot Classification.} In this setting, train and test categories coincide, and we evaluate performance across different numbers of shots ($K=1,2,4,8,16$) and different CLIP backbones (ViT-B/16 and ViT-B/32) on the same 11 datasets.

\textbf{Implementation Details.} We use a ViT-B/16-based CLIP model as the default backbone and report results averaged over 3 seeds. We adopt deep prompting with $T = V = 4$ prompt tokens and parameterize prompts through a rank-1 factorization ($r = 1$) with the shared up-projection. For domain generalization, we train the ImageNet source model on all classes with $K = 16$ shots in the first 3 transformer layers. For few-shot and base-to-novel settings, prompts are learned in the first 9 transformer layers. Low-rank prompts are randomly initialized with a normal distribution with standard deviation $0.05$. The consistency loss weights are set to $\lambda_1 = 25$ and $\lambda_2 = 10$ for $\mathcal{L}_{\text{SCL-text}}$ and $\mathcal{L}_{\text{SCL-image}}$, respectively. We train for 30 epochs for base-to-novel and 50 epochs for the few-shot and domain generalization settings, using SGD with a learning rate of 0.0025. All hyperparameters are fixed across all datasets and benchmarks. For the zero-shot text features used in $\mathcal{L}_{\text{SCL}}$, we use an ensemble of $N = 60$ standard text templates provided in~\cite{radford2021learning}. All experiments are conducted on four NVIDIA A6000 GPUs.

\subsection{Main Results}

\textbf{Base-to-Novel Generalization.}
Table~\ref{tab:base_to_novel} reports results across 11 datasets under the base-to-novel
generalization protocol, where models are trained on base classes with 16 shots and
evaluated on both base and novel classes. MMLoP achieves an average harmonic mean
of \textbf{79.70\%}, outperforming a wide range of recent and more sophisticated
methods, including PromptSRC (79.62\%, 46K parameters), TCP (79.51\%, 332K parameters), MetaPrompt (79.09\%, 31K parameters),
LoGoPrompt (79.03\%), LASP (78.61\%), MaPLe (78.55\%, 3.55M parameters),
ProVP (78.76\%, 147K parameters), RPO (77.78\%), and CLIP-LoRA (77.28\%,
184K parameters). MMLoP also remains competitive with MMA (79.87\%, 581K
parameters), 2SFS (80.20\%, 170K parameters), and CoPrompt (80.48\%, 3.82M
parameters) --- methods that require \textbf{51$\times$}, \textbf{15$\times$}, and
\textbf{332$\times$} more trainable parameters, respectively. With only
\textbf{11.5K trainable parameters}, MMLoP operates at a scale comparable to
early text-only methods like CoOp (8.2K), while benefiting from deep multi-modal
prompting across both encoders. Notably, MMLoP demonstrates strong novel class
accuracy of \textbf{75.98\%} on average, reflecting a $+$4.19\% gain over the IVLP
baseline and confirming that our regularization components effectively prevent
overfitting to base classes. On Food101 it achieves a novel accuracy of 91.70\%,
and on EuroSAT a harmonic mean of 85.22\%, substantially surpassing MaPLe
(82.35\%) and CoCoOp (71.21\%). The performance gap relative to the top-performing methods is most apparent on fine-grained datasets such as FGVCAircraft, where
discriminative visual features are harder to capture within a low-rank subspace. Overall, these results show that MMLoP lies on the \textit{accuracy--efficiency Pareto
frontier}: every compared method with higher HM uses substantially more parameters,
and every method with fewer parameters achieves lower HM. MMLoP thus outperforms
the majority of existing methods while using a fraction of their parameter budgets. We further compare against dedicated low-rank adaptation methods in Appendix~C.

\begin{table}[t]
\centering
\caption{Domain generalization performance on ImageNet variants.}
\label{tab:dg}
\renewcommand\arraystretch{1.2}
\resizebox{\textwidth}{!}{%
    \footnotesize
    \begin{tabular}{lcccccccccccc}
    \toprule
    \textbf{Dataset} 
    & CLIP~\cite{radford2021learning} 
    & UPT~\cite{zang2022unified}
    & CoOp~\cite{COOP}
    & CoCoOp~\cite{cocoop} 
    & ProGrad~\cite{zhu2023prompt} 
    & KgCoOp~\cite{yao2023visual} 
    & TaskRes~\cite{yu2023task} 
    & MaPLe~\cite{khattak2023maple} 
    & RPO~\cite{lee2023read} 
    & MMA~\cite{yang2024mma} 
    & CoPrompt~\cite{royconsistency} 
    & \textbf{MMLoP} \\ 
    \midrule
    ImageNet  & 66.73 & 72.63 & 71.51 & 71.02 & 72.24 & 71.20 & {73.90} & 70.72 & 71.67 & 71.00 & 70.80 & 71.00 \\ 
    ImageNet-V2      & 60.83 & 64.35 & 64.20 & 64.07 & 64.73 & 64.10 & {65.85} & 64.07 & 65.13 & 64.33 & 64.25 & 64.30 \\ 
    ImageNet-S       & 46.15 & 48.66 & 47.99 & 48.75 & 47.61 & 48.97 & 47.70 & 49.15 & 49.27 & 49.13 & {49.43} & 49.07 \\ 
    ImageNet-A       & 47.77 & 50.66 & 49.71 & 50.63 & 49.39 & 50.69 & 49.17 & 50.90 & 50.13 & {51.12} & 50.50 & 50.83 \\ 
    ImageNet-R       & 73.96 & 76.24 & 75.21 & 76.18 & 74.58 & 76.70 & 75.23 & 76.98 & 76.57 & 77.32 & 77.51 & {77.63} \\ 
    \midrule
    \textbf{Average} & 57.18 & 59.98 & 59.28 & 59.91 & 59.07 & 60.11 & 59.49 & 60.28 & 60.27 & 60.48 & 60.42 & {60.46} \\ 
    \bottomrule
    \end{tabular}%
}
\end{table}

\textbf{Domain Generalization.}
Table~\ref{tab:dg} evaluates robustness to distribution shift by training on ImageNet
and directly evaluating on four out-of-distribution variants: ImageNet-V2,
ImageNet-Sketch, ImageNet-A, and ImageNet-R. MMLoP achieves a competitive
average target accuracy of \textbf{60.46\%}, outperforming the majority of compared
methods including MaPLe (60.28\%), CoPrompt (60.42\%), and RPO (60.27\%),
while using only 11.5K trainable parameters. Notably, MMLoP attains the highest
accuracy on ImageNet-R (\textbf{77.63\%}) among all methods, suggesting that our
consistency regularization and low-rank parameterization effectively preserve
CLIP's pretrained representations and prevent source-domain overfitting. While MMA
(60.48\%) achieves a marginally higher average target accuracy by \textbf{0.02\%},
it requires \textbf{51$\times$} more trainable parameters. These results demonstrate
that MMLoP generalizes robustly across domain shifts without sacrificing parameter
efficiency.

\begin{table*}[t]
    \def\arraystretch{1.1}
    \centering
    \footnotesize
    \caption{\emph{All-to-all} experiments, where train/test categories coincide, with the ViT-B/16 backbone, using $k=4,8,16$ shots per class.}
\resizebox{0.8\textwidth}{!}{
    \begin{tabular}{llccccccccccc|c}
    \toprule
     {\textbf{Shots}} &  {\textbf{Method}} &  {\textbf{ImageNet}} &  {\textbf{SUN}} &  {\textbf{AIR}} &  {\textbf{ESAT}} &  {\textbf{CARS}} &  {\textbf{FOOD}} &  {\textbf{PETS}} &  {\textbf{FLWR}} &  {\textbf{CAL}} &  {\textbf{DTD}} &  {\textbf{UCF}} &  {\textbf{Mean}} \\
    \midrule
    
    \cellcolor{gray!0} \multirow{13}{*}{16} 
& \emph{Zero-Shot} &  66.7 & 62.6 & 24.7 & 47.5 & 65.3 & 86.1 & 89.1 & 71.4 & 92.9 & 43.6 & 66.7 & 65.1 \\

& CoOp \cite{COOP} (ctx=16) &  71.9 & 74.9 & 43.2 & 85.0 & 82.9 & 84.2 & 92.0 & 96.8 & 95.8 & 69.7 & 83.1 & 80.0 \\

& CoCoOp \cite{cocoop} &  71.1 & 72.6 & 33.3 & 73.6 & 72.3 & {87.4} & 93.4 & 89.1 & 95.1 & 63.7 & 77.2 & 75.4 \\

& TIP-Adapter-F \cite{zhang2021tip} &  73.4 & 76.0 & 44.6 & 85.9 & 82.3 & 86.8 & 92.6 & 96.2 & 95.7 & 70.8 & 83.9 & 80.7 \\

& CLIP-Adapter \cite{gao2024clip} &  69.8 & 74.2 & 34.2 & 71.4 & 74.0 & 87.1 & 92.3 & 92.9 & 94.9 & 59.4 & 80.2 & 75.5 \\


& KgCoOp \cite{yao2023visual} &  70.4 & 73.3 & 36.5 & 76.2 & 74.8 & 87.2 & 93.2 & 93.4 & 95.2 & 68.7 & 81.7 & 77.3 \\

& TaskRes \cite{yu2023task} &  73.0 & 76.1 & 44.9 & 82.7 & 83.5 & 86.9 & 92.4 & 97.5 & 95.8 & 71.5 & 84.0 & 80.8 \\

& MaPLe \cite{khattak2023maple} &  71.9 & 74.5 & 36.8 & 87.5 & 74.3 & {87.4} & 93.2 & 94.2 & 95.4 & 68.4 & 81.4 & 78.6 \\

& ProGrad \cite{zhu2023prompt} &  72.1 & 75.1 & 43.0 & 83.6 & 82.9 & 85.8 & 92.8 & 96.6 & 95.9 & 68.8 & 82.7 & 79.9 \\

& LP++ \cite{huang2024lp++} & 73.0 & 76.0 & 42.1 & 85.5 & 80.8 & 87.2 & 92.6 & 96.3 & 95.8 & 71.9 & 83.9 & 80.5 \\
& CLIP-LoRA \cite{zanella2024low} & {73.6} & 76.1 & {54.7} & {92.1} & {86.3} & 84.2 & 92.4 & {98.0} & {96.4} & 72.0 & {86.7} & {83.0} \\
& MMA \cite{yang2024mma} & 73.2 & {76.6} & 44.7 & 85.0 & 80.2 & 87.0 & {93.9} & 96.8 & 95.8 & {72.7} & 85.0 & 81.0 \\


\rowcolor{lightgray!30} \cellcolor{white} &
 MMLoP & 71.9 & 75.8 & 45.5 & 92.2 & 80.4 & 87.4 & 93.9 & 97.2 & 95.7 & 72.6 & 84.3 & {81.5}\\ 
    \cmidrule(lr){1-14}
    
    \cellcolor{gray!0} \multirow{13}{*}{8} & \emph{Zero-Shot} & 66.7 & 62.6 & 24.7 & 47.5 & 65.3 & 86.1 & 89.1 & 71.4 & 92.9 & 43.6 & 66.7 & 65.1 \\
 & CoOp \cite{COOP} (ctx=16) & 70.6 & 71.9 & 38.5 & 77.1 & 79.0 & 82.7 & 91.3 & 94.9 & 94.5 & 64.8 & 80.0 & 76.8 \\
 & CoCoOp \cite{cocoop} & 70.8 & 71.5 & 32.4 & 69.1 & 70.4 & {87.0} & {93.3} & 86.3 & 94.9 & 60.1 & 75.9 & 73.8 \\
 & TIP-Adapter-F \cite{zhang2021tip} & 71.7 & 73.5 & 39.5 & 81.3 & 78.3 & 86.9 & 91.8 & 94.3 & 95.2 & 66.7 & 82.0 & 78.3 \\
 & CLIP-Adapter \cite{gao2024clip} & 69.1 & 71.7 & 30.5 & 61.6 & 70.7 & 86.9 & 91.9 & 83.3 & 94.5 & 50.5 & 76.2 & 71.5 \\
 & KgCoOp \cite{yao2023visual} & 70.2 & 72.6 & 34.8 & 73.9 & 72.8 & {87.0} & 93.0 & 91.5 & 95.1 & 65.6 & 80.0 & 76.0 \\
 & TaskRes \cite{yu2023task} & {72.3} & 74.6 & 40.3 & 77.5 & 79.6 & 86.4 & 92.0 & {96.0} & 95.3 & 66.7 & 81.6 & 78.4 \\
 & MaPLe \cite{khattak2023maple} & 71.3 & 73.2 & 33.8 & 82.8 & 71.3 & {87.2} & {93.1} & 90.5 & 95.1 & 63.0 & 79.5 & 76.4 \\
 & ProGrad \cite{zhu2023prompt} & 71.3 & 73.0 & 37.7 & 77.8 & 78.7 & 86.1 & 92.2 & 95.0 & 94.8 & 63.9 & 80.5 & 77.4 \\
 & LP++ \cite{huang2024lp++} & 72.1 & {75.1} & 39.0 & 78.2 & 76.4 & 86.8 & 91.8 & 95.2 & 95.5 & {67.7} & 81.9 & 78.2 \\
 & CLIP-LoRA \cite{zanella2024low} & {72.3} & 74.7 & {45.7} & {89.7} & {82.1} & 83.1 & 91.7 & {96.3} & {95.6} & 67.5 & {84.1} & {80.3} \\
 & MMA \cite{yang2024mma} & 71.9 & 74.7 & 38.9 & 69.7 & 76.8 & 86.4 & 92.9 & 94.6 & {95.6} & 66.9 & 82.9 & 77.4 \\

\rowcolor{lightgray!30} \cellcolor{white} &
 MMLoP & 71.5 & 74.7 & 40.7 & 88.2 & 77.8 & 87.0 & 93.3 & 95.8 & 95.3 & 68.9 & 82.7 & {79.6}\\  
    \cmidrule(lr){1-14}
    
    \cellcolor{gray!0} \multirow{13}{*}{4} & \emph{Zero-Shot}  & 66.7 & 62.6 & 24.7 & 47.5 & 65.3 & 86.1 & 89.1 & 71.4 & 92.9 & 43.6 & 66.7 & 65.1 \\
 & CoOp \cite{COOP} (ctx=16) & 68.8 & 69.7 & 30.9 & 69.7 & 74.4 & 84.5 & 92.5 & 92.2 & 94.5 & 59.5 & 77.6 & 74.0 \\
 & CoCoOp \cite{cocoop} & 70.6 & 70.4 & 30.6 & 61.7 & 69.5 & 86.3 & 92.7 & 81.5 & 94.8 & 55.7 & 75.3 & 71.7 \\
 & TIP-Adapter-F \cite{zhang2021tip} & 70.7 & 70.8 & 35.7 & 76.8 & 74.1 & 86.5 & 91.9 & 92.1 & 94.8 & 59.8 & 78.1 & 75.6 \\
 & CLIP-Adapter \cite{gao2024clip} & 68.6 & 68.0 & 27.9 & 51.2 & 67.5 & 86.5 & 90.8 & 73.1 & 94.0 & 46.1 & 70.6 & 67.7 \\
 & KgCoOp \cite{yao2023visual} & 69.9 & 71.5 & 32.2 & 71.8 & 69.5 & {86.9} & 92.6 & 87.0 & 95.0 & 58.7 & 77.6 & 73.9 \\
 & TaskRes \cite{yu2023task} & 71.0 & 72.7 & 33.4 & 74.2 & 76.0 & 86.0 & 91.9 & 85.0 & 95.0 & 60.1 & 76.2 & 74.7 \\
 & MaPLe \cite{khattak2023maple} & 70.6 & 71.4 & 30.1 & 69.9 & 70.1 & {86.7} & {93.3} & 84.9 & 95.0 & 59.0 & 77.1 & 73.5 \\
 & ProGrad \cite{zhu2023prompt} & 70.2 & 71.7 & 34.1 & 69.6 & 75.0 & 85.4 & 92.1 & 91.1 & 94.4 & 59.7 & 77.9 & 74.7 \\
 & LP++ \cite{huang2024lp++} & 70.8 & {73.2} & 34.0 & 73.6 & 74.0 & 85.9 & 90.9 & 93.0 & 95.1 & 62.4 & 79.2 & 75.6 \\
 & CLIP-LoRA \cite{zanella2024low} & {71.4} & 72.8 & {37.9} & {84.9} & {77.4} & 82.7 & 91.0 & {93.7} & {95.2} & {63.8} & {81.1} & {77.4} \\
 & MMA \cite{yang2024mma} & 70.5 & 72.9 & 35.0 & 42.4 & 73.3 & 86.0 & {92.9} & 91.3 & 94.5 & 60.1 & 79.0 & 72.5 \\

\rowcolor{lightgray!30} \cellcolor{white} & MMLoP & 70.8 & 73.1 & 36.1 & 84.5 & 74.9 & 86.1 & 93.2 & 93.3 & 95.2 & 64.7 & 80.2 & {77.5}
\\
 \cmidrule(lr){1-14}
    \end{tabular}
    
\label{tab:all2all}
}
\end{table*}

\textbf{All-to-All Few-Shot Classification.}
Table~\ref{tab:all2all} reports all-to-all few-shot results on 11 datasets using the
ViT-B/16 backbone across $K \in \{4, 8, 16\}$ shots, where train and test categories
coincide. MMLoP demonstrates consistently strong performance across all shot
settings, achieving mean accuracies of \textbf{81.5\%}, \textbf{79.6\%}, and
\textbf{77.5\%} for 16, 8, and 4 shots, respectively. At 16 shots, MMLoP ranks
second overall, surpassing MMA (81.0\%), MaPLe (78.6\%), and all non-adapter
baselines, while remaining competitive with CLIP-LoRA (83.0\%) which uses
dedicated LoRA adapters on the full backbone rather than prompt tokens. At 4
shots, MMLoP achieves the \textbf{highest mean accuracy of 77.5\%} among all
compared methods, outperforming CLIP-LoRA (77.4\%) and LP++ (75.6\%),
highlighting the strength of our low-rank parameterization and consistency
regularization in the extremely low-data regime. Notably, MMLoP achieves
particularly strong results on EuroSAT across all shot settings (92.2\%, and
84.5\% for 16 and 4 shots respectively), demonstrating robust adaptation to
specialized domains. These results validate that MMLoP achieves strong
few-shot adaptation across diverse datasets and data regimes, with only 11.5K
trainable parameters. Additional results using the ViT-B/32 backbone are reported in 
Table~A1. Few-shot curves for both ViT-B/16 and ViT-B/32 
are visualized in Figs.~A2 and~A3 (Appendix).

\subsection{Ablation Study}

Table~\ref{tab:ablations_on_components} validates the contribution of each proposed component
through an incremental ablation study averaged over 11 datasets. Starting from
the IVLP baseline (77.51\% HM), adding the low-rank factorization alone
reduces both base and novel accuracy (77.06\% HM), as the rank-1 constraint
limits the prompt's expressive capacity without any compensating regularization.
Incorporating the Self-Regulating Consistency Loss ($\mathcal{L}_{\text{SCL}}$)
yields a substantial gain in novel class accuracy from 71.39\% to 74.48\%,
boosting the harmonic mean to 78.78\% --- confirming that explicit feature-level
anchoring to the frozen CLIP representations is critical for generalization to
unseen classes. Adding Uniform Drift Correction (UDC) further improves novel
accuracy to 75.12\% (HM: 79.20\%) by removing the global embedding shift shared
across all class embeddings. Finally, the Shared Up-Projection pushes novel
accuracy to \textbf{75.98\%} and the harmonic mean to \textbf{79.70\%} by
coupling vision and text prompts through a common low-rank factor, enforcing
cross-modal alignment at virtually no additional parameter cost. We note that these gains in novel class generalization (+4.19\% over IVLP) 
come at the cost of a modest reduction in base class accuracy ($-$0.42\%), 
which is expected given that our regularization components explicitly 
discourage the model from over-specializing to the base training classes 
in favor of preserving CLIP's broader representational capacity. Per-dataset results are in Table~A4 
in Appendix.


\begin{table}[!t]
\caption{Effect of our proposed regularization techniques. Results are averaged over 11 datasets. }
\label{tab:ablations_on_components}
\small \centering
\renewcommand\arraystretch{1.2}
\setlength{\tabcolsep}{8pt}
\scalebox{0.7}[0.7]{
\begin{tabular}{|l|cc|c|}
\hline
\textbf{Method} & \textbf{Base Acc.} & \textbf{Novel Acc.} & \textbf{HM} \\ \hline
1: Independent V-L prompting & \textbf{84.21} & 71.79 & 77.51 \\ 
2: + low-rank prompting & 83.70 & 71.39 & 77.06 \\
3: + $\mathcal{L}_\text{SCL}$ & 83.60 & 74.48 & 78.78 \\ 
4: + UDC & 83.78 & 75.12 & 79.20 \\ 
\rowcolor{lightgray!30}
5: + Shared Up-Projection & 83.79 & \textbf{75.98} & \textbf{79.70} \\ \hline
\end{tabular}
}
\end{table}

\begin{table*}[t]
\caption{Hyperparameter sensitivity analysis. Results are averaged over 9 datasets. }
\label{tab:hyperparams}
\small \centering
\renewcommand\arraystretch{1.2}

\begin{subtable}[t]{0.3\textwidth}
\centering
\label{tab:prompt_depth}
\scalebox{0.85}[0.85]{
\begin{tabular}{|c|ccc|}
\hline
\textbf{Depth} & \textbf{Base} & \textbf{Novel} & \textbf{HM} \\ \hline
7  & 83.63 & 74.19 & 78.63 \\
9  & 84.44 & \textbf{75.85} & \textbf{79.91} \\
11 & \textbf{84.87} & 74.58 & 79.39 \\ \hline
\end{tabular}
}
\end{subtable}
\hfill
\begin{subtable}[t]{0.3\textwidth}
\centering
\label{tab:prompt_length}
\scalebox{0.85}[0.85]{
\begin{tabular}{|c|ccc|}
\hline
\textbf{Length} & \textbf{Base} & \textbf{Novel} & \textbf{HM} \\ \hline
2 & 84.16 & 74.99 & 79.31 \\
4  & 84.44 & \textbf{75.85} & \textbf{79.91} \\
8 & \textbf{84.83} & 74.38 & 79.26 \\ \hline
\end{tabular}
}
\end{subtable}
\hfill
\begin{subtable}[t]{0.3\textwidth}
\centering
\label{tab:lora_rank}
\scalebox{0.85}[0.85]{
\begin{tabular}{|c|ccc|}
\hline
\textbf{Rank} & \textbf{Base} & \textbf{Novel} & \textbf{HM} \\ \hline
1 & 84.44 & \textbf{75.85} & \textbf{79.91} \\
2 & 84.76 & 75.01 & 79.59 \\
4 & \textbf{84.80} & 75.20 & 79.71 \\ \hline
\end{tabular}
}
\end{subtable}

\end{table*}

\begin{table}[t]
  \centering
  \caption{Efficiency comparison averaged over 11 datasets (ViT-B/16, single A6000).}
  \label{tab:efficiency}
  \resizebox{0.8\linewidth}{!}{%
  \begin{tabular}{lcccccccc}
    \hline\noalign{\smallskip}
    Method & Params (M) & VRAM (GB) & Train (min) & Train (ms/img) & Infer (ms/img) & FPS & GFLOP & HM \\
    \noalign{\smallskip}\hline\noalign{\smallskip}
    CoOp~\cite{COOP}            & 0.008 & 1.96 & 58.1 & 19.39 & 1.94 & 561.6 & 324.6 & 71.66 \\
    PromptSRC~\cite{khattak2023self}       & 0.046 & 2.99 & 44.7 & 30.16 & 2.29 & 530.0 & 324.9 & 79.62 \\
    MaPLe~\cite{khattak2023maple}           & 3.555 & 2.27 &  7.8 & 19.75 & 2.03 & 545.4 & 324.7 & 78.55 \\
    MMA~\cite{yang2024mma}             & 0.581 & 2.57 &  7.9 & 25.84 & 2.39 & 489.6 & 326.5 & 79.87 \\
    CoPrompt~\cite{royconsistency}        & 3.818 & 2.62 &  9.6 & 20.49 & 1.39 & 791.7 & 342.3 & 80.48 \\
    CLIP-LoRA~\cite{zanella2024low}       & 0.184 & 4.39 & 41.4 & 34.32 & 2.27 & 487.8 & 324.7 & 77.28 \\
    Comp-LoRA~\cite{wang2025complementary}       & 0.184 & 4.62 & 41.6 & 33.20 & 1.83 & 552.0 & 346.2 & 78.77 \\
    Block-LoRA~\cite{zhou2025one}      & 0.230 & 4.39 & 56.4 & 33.53 & 1.82 & 555.5 & 324.7 & 78.82 \\
    \rowcolor{gray!15}
    MMLoP (Ours)         & 0.012 & 2.33 & 24.7 & 20.28 & 1.79 & 638.0 & 324.9 & 79.70 \\
    \noalign{\smallskip}\hline
  \end{tabular}%
  }
\end{table}

\subsection{Hyperparameter Sensitivity}\label{hyper_sens}
Table~\ref{tab:hyperparams} analyzes the sensitivity of MMLoP to three key
hyperparameters: prompt depth, prompt length, and prompt factorization rank. For prompt depth, performance peaks at depth 9 (HM: 79.91\%), with shallower prompting (depth 7) yielding a notable drop to 78.63\%, while deeper prompting (depth 11) improves base accuracy (84.87\%) but hurts novel accuracy (74.58\%), suggesting overfitting
to base classes when prompts are applied too deeply. For prompt length, depth 4 achieves the best harmonic mean (79.91\%), with both shorter (length 2: 79.31\%) and longer (length 8: 79.26\%) configurations performing slightly worse, indicating that the model is not highly sensitive to this parameter around the chosen value.
For prompt factorization rank, rank 1 achieves the best harmonic mean (79.91\%) and the highest
novel accuracy (75.85\%), with higher ranks (2 and 4) improving base accuracy
marginally but consistently hurting novel generalization --- confirming that the stronger implicit regularization of a rank-1 factorization is beneficial for
generalization to unseen classes. Overall, MMLoP is reasonably robust to
hyperparameter choices, and all hyperparameters are fixed across all datasets and benchmarks without any dataset-specific tuning. Detailed per-dataset results are provided in Tables~A5, A6, and~A7 in the appendix.

\subsection{Efficiency Analysis}
Tab.~\ref{tab:efficiency} reports trainable parameters, peak VRAM, training and inference cost, and HM averaged over 11 datasets (ViT-B/16, single A6000). MMLoP achieves the lowest parameter count (0.012M, $\sim$50--330$\times$ fewer than multi-modal baselines) while remaining competitive across all system-level metrics: VRAM (2.33 GB) is lower than 6 of 8 baselines; per-iteration training cost (20.28 ms/img) matches CoOp/MaPLe and is $\sim$40\% faster than the LoRA family. The lower total training time (Train min) of MaPLe, MMA, and CoPrompt reflects their use of very few epochs (5--8) to avoid overfitting, not per-iteration efficiency. Inference latency (1.79 ms/img) is faster than CoOp, MaPLe, PromptSRC, MMA, and CLIP-LoRA, demonstrating that UDC and the consistency loss add no inference-time overhead.

\section{Conclusion}

We presented MMLoP, a parameter-efficient multi-modal prompt learning framework
that achieves deep vision-language prompting with only 11.5K trainable parameters.
By parameterizing prompts through a low-rank factorization with a shared
up-projection, and combining this with a self-regulating consistency loss and
uniform drift correction, MMLoP enforces cross-modal alignment while preventing
overfitting to base classes and preserving CLIP's pretrained representations.
Extensive experiments across base-to-novel generalization, domain generalization,
and few-shot classification show that MMLoP outperforms most
existing methods at a fraction of their parameter budgets.



\section*{Acknowledgments}

This work was supported by the National Science Foundation under Grant 2419982,
Grant 2342253, and Grant 2236483.
\clearpage

%
%
\bibliographystyle{splncs04}
\bibliography{main}

\clearpage
\appendix
\begin{center}
    {\Large\bfseries Supplementary Material:\\MMLoP: Multi-Modal Low-Rank Prompting for Efficient Vision-Language Adaptation\par}
    \vspace{7em}
\end{center}
\renewcommand{\thefigure}{A\arabic{figure}}
\renewcommand{\thetable}{A\arabic{table}}
\setcounter{figure}{0}
\setcounter{table}{0}

\section{Analysis of Learned Shared Up-Projection}
\begin{figure}[t]
  \centering
\includegraphics[width=1.0\textwidth]{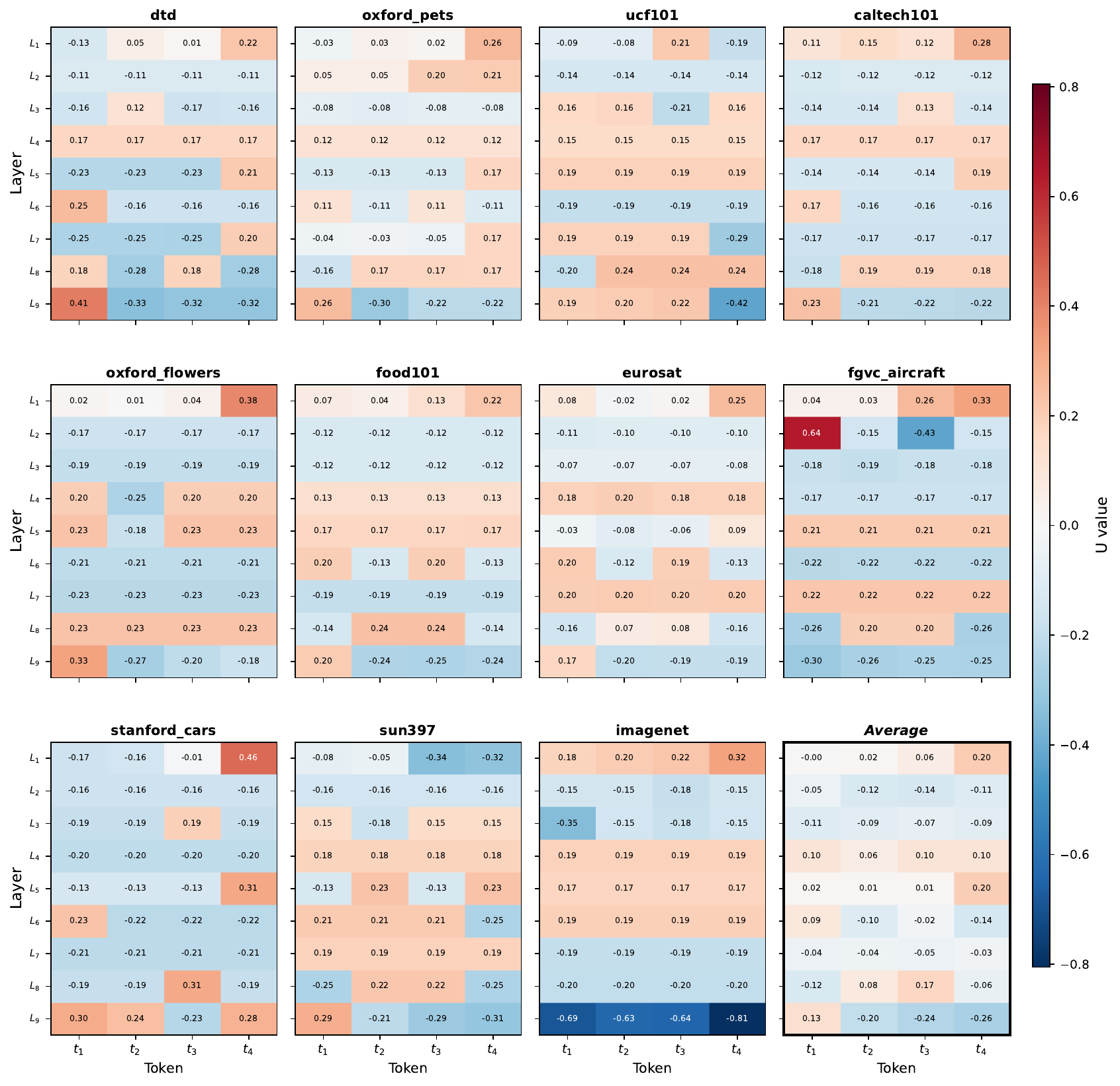}
  \caption{Visualization of the learned shared up-projection matrix $\bm{U}^{(l)}$ 
across transformer layers and prompt tokens for each of the 11 datasets and their 
average.}
  \label{u_heatmap_comparison}
\end{figure}

Figure~\ref{u_heatmap_comparison} visualizes the learned values of the shared 
up-projection matrix $\bm{U}^{(l)}$ across transformer layers and prompt tokens 
for each of the 11 datasets. The final layer ($L_9$) consistently exhibits the 
largest magnitude activations across all datasets, suggesting that the model 
relies most heavily on prompt-induced feature modulation at the deepest layers 
where task-specific representations are formed. Furthermore, the four prompt 
tokens $t_1$--$t_4$ learn meaningfully different activation patterns across many 
layers, indicating that each token captures distinct aspects of the task-specific 
adaptation despite sharing the same up-projection structure.

\section{Additional All-to-All Few-Shot Experiments}

\begin{figure}[t]
  \centering
\includegraphics[width=1\textwidth]{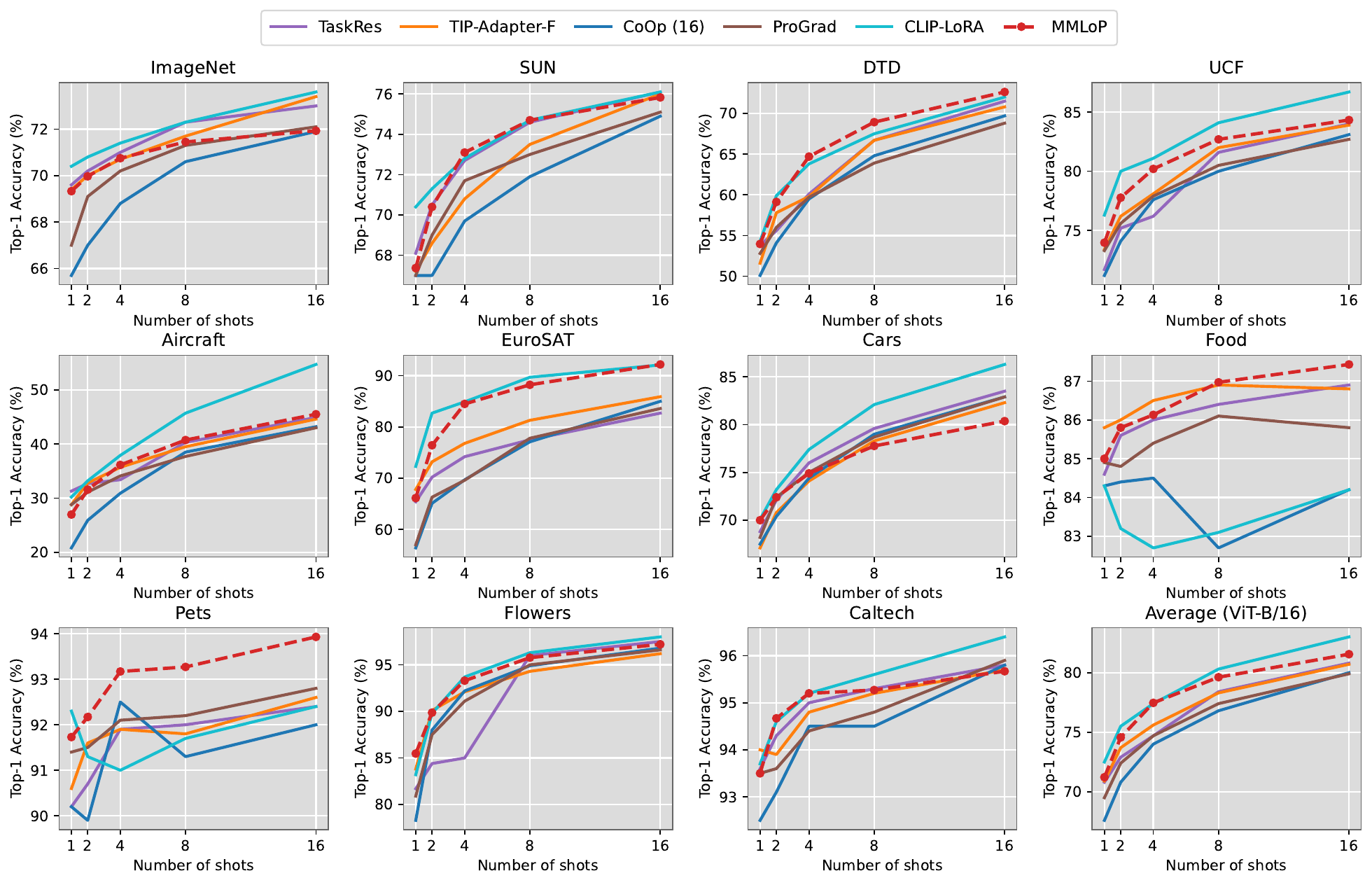}
  \caption{All-to-all few-shot classification results on 11 datasets using the 
ViT-B/16 backbone across $K \in \{1, 2, 4, 8, 16\}$ shots.}
  \label{vit16}
\end{figure}

\begin{figure}[t]
  \centering
\includegraphics[width=1\textwidth]{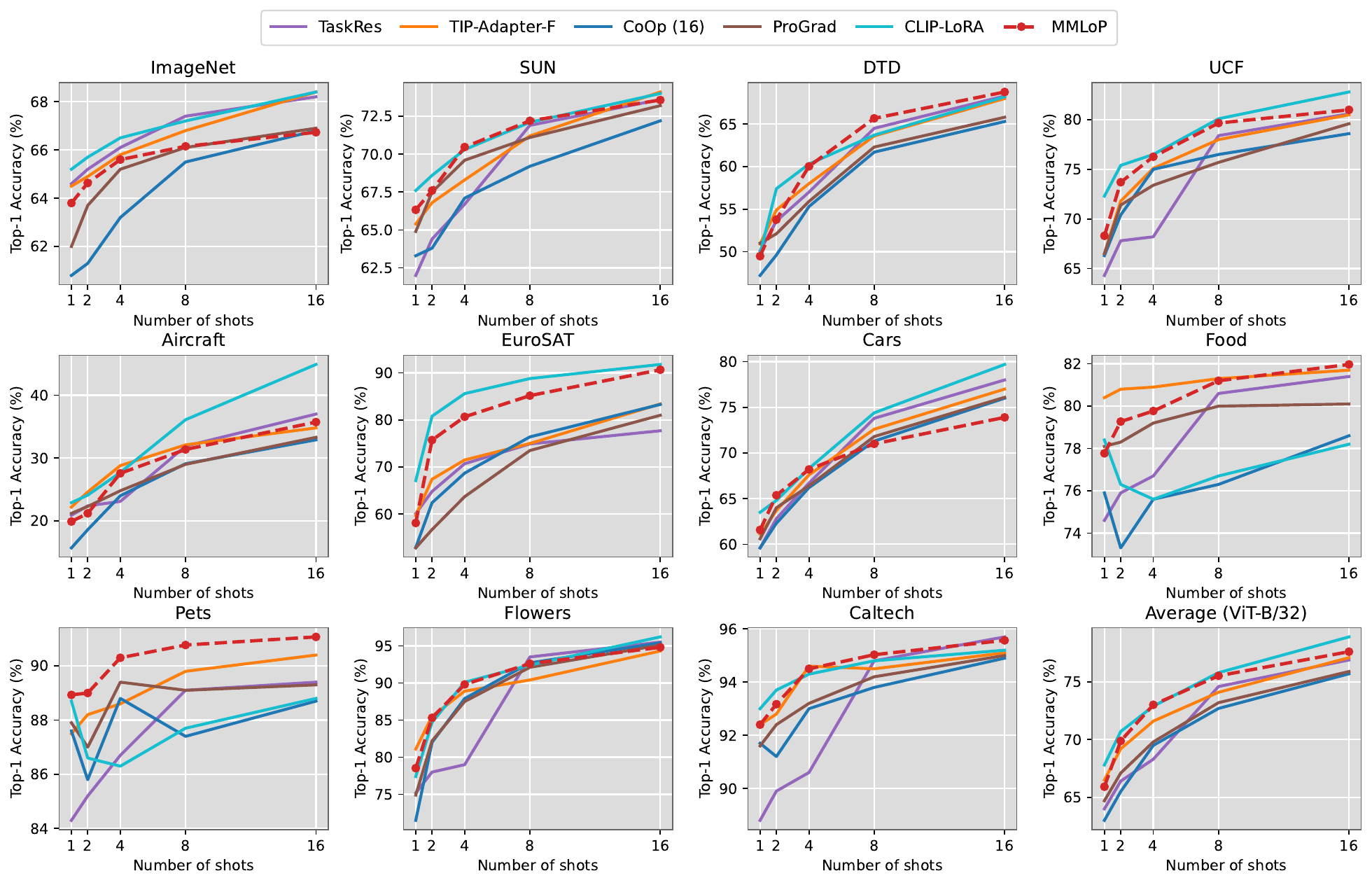}
  \caption{All-to-all few-shot classification results on 11 datasets using the 
ViT-B/32 backbone across $K \in \{1, 2, 4, 8, 16\}$ shots.}
  \label{vit32}
\end{figure}

Table~\ref{tab:vit32} reports all-to-all few-shot classification 
results using the ViT-B/32 backbone across $K \in \{4, 8, 16\}$ shots 
on 11 datasets, and Figures~\ref{vit16} 
and~\ref{vit32} visualize the full learning curves across 
$K \in \{1, 2, 4, 8, 16\}$ shots for both ViT-B/16 and ViT-B/32 
backbones. MMLoP achieves competitive average accuracies across both 
backbones and all shot settings, with particularly strong performance 
in the low-data regime. At $K = 16$, MMLoP ranks second overall on 
ViT-B/16 and remains competitive on ViT-B/32, while at $K \leq 8$ it 
consistently matches or outperforms the majority of compared methods on 
average, highlighting the effectiveness of the low-rank parameterization 
and consistency regularization when training data is scarce. These 
results confirm that MMLoP generalizes well across different backbone 
scales without any backbone-specific tuning.

\begin{table*}
    \def\arraystretch{1.1}
    \centering
    \small
    \caption{\emph{All-to-all} experiments, where train/test categories coincide, with the ViT-B/32 backbone, using $k=4,8,16$ shots per class.}
    \begin{adjustbox}{max width=\textwidth}
    \begin{tabular}{llccccccccccc|c}
    \toprule
     \textsc{\textbf{Shots}} &  \textsc{\textbf{Method}} &  \textsc{\textbf{ImageNet}} &  \textsc{\textbf{SUN}} &  \textsc{\textbf{AIR}} &  \textsc{\textbf{ESAT}} &  \textsc{\textbf{CARS}} &  \textsc{\textbf{FOOD}} &  \textsc{\textbf{PETS}} &  \textsc{\textbf{FLWR}} &  \textsc{\textbf{CAL}} &  \textsc{\textbf{DTD}} &  \textsc{\textbf{UCF}} &  \textsc{\textbf{Mean}} \\
    \midrule
    
    \cellcolor{gray!0}\multirow{13}{*}{16} 
& \emph{Zero-Shot} & 61.9 & 62.0 & 19.3 & 45.1 & 60.4 & 80.5 & 87.5 & 67.0 & 91.1 & 42.6 & 62.2 & 61.8 \\

& CoOp \cite{COOP} (ctx=16) & 66.8 & 72.2 & 32.9 & 83.3 & 76.0 & 78.6 & 88.7 & 95.4 & 94.9 & 65.3 & 78.6 & 75.7 \\

& CoCoOp \cite{cocoop} & 66.0 & 69.8 & 22.6 & 70.4 & 64.6 & {81.9} & {91.0} & 82.5 & 94.3 & 59.7 & 75.3 & 70.7 \\

& TIP-Adapter-F \cite{zhang2021tip} & {68.4} & {74.1} & 34.8 & 83.4 & 77.0 & 81.7 & 90.4 & 94.3 & 95.1 & 68.0 & 80.5 & 77.1 \\

& CLIP-Adapter \cite{gao2024clip} & 64.9 & 71.8 & 26.7 & 64.7 & 68.9 & {81.9} & 90.1 & 88.7 & 94.8 & 58.1 & 76.5 & 71.6 \\


& KgCoOp \cite{yao2023visual} & 65.4 & 71.0 & 23.7 & 70.1 & 67.3 & 81.7 & 90.8 & 86.1 & 94.4 & 65.1 & 77.5 & 72.1 \\

& TaskRes  \cite{yu2023task} & 68.2 & 73.6 & 37.0 & 77.7 & 78.0 & 81.4 & 89.4 & 95.5 & 95.7 & 68.3 & 80.6 & 76.9 \\

& MaPLe \cite{khattak2023maple} & 66.7 & 72.0 & 28.0 & 83.3 & 66.9 & 82.1 & 91.7 & 89.0 & 95.1 & 63.4 & 77.3 & 74.1 \\

& ProGrad \cite{zhu2023prompt} & 66.9 & 73.2 & 33.3 & 81.0 & 76.1 & 80.1 & 89.3 & 95.1 & 95.0 & 65.8 & 79.6 & 75.9 \\

& LP++ \cite{huang2024lp++} & 68.1 & 74.0 & 34.3 & 82.8 & 75.2 & 81.8 & 90.5 & 93.9 & 95.0 & 67.8 & 80.1 & 76.7 \\

& CLIP-LoRA \cite{zanella2024low} & {68.4} & 74.0 & {44.9} & {91.8} & {79.7} & 78.2 & 88.8 & 96.2 & 95.2 & 68.2 & {82.8} & {{78.9}} \\
& MMA \cite{yang2024mma} & 68.0 & 74.0 & 34.0 & 80.1 & 73.5 & 81.4 & {91.5} & 94.3 & {95.6} & {68.9} & 81.7 & 76.7 \\


\rowcolor{lightgray!30} \cellcolor{white} &
 MMLoP & 66.8 & 73.6 & 35.7 & 90.7 & 73.9 & 82.0 & 91.1 & 94.8 & 95.6 & 68.8 & 81.0 & {77.6}\\
 \\ [-5ex] \\ 
    \cmidrule(lr){1-14}
    \cellcolor{gray!0} \multirow{13}{*}{8} & \emph{Zero-Shot}  & 61.9 & 62.0 & 19.3 & 45.1 & 60.4 & 80.5 & 87.5 & 67.0 & 91.1 & 42.6 & 62.2 & 61.8 \\
 & CoOp \cite{COOP} (ctx=16) & 65.5 & 69.2 & 29.1 & 76.4 & 71.3 & 76.3 & 87.4 & 92.7 & 93.8 & 61.7 & 76.5 & 72.7 \\
 & CoCoOp \cite{cocoop} & 65.8 & 68.9 & 20.3 & 58.1 & 63.4 & 81.6 & 90.1 & 77.3 & 93.8 & 57.4 & 72.4 & 68.1 \\
 & TIP-Adapter-F \cite{zhang2021tip} & 66.8 & 71.2 & 32.1 & 75.0 & 72.6 & 81.3 & 89.8 & 90.4 & 94.5 & 63.6 & 78.0 & 74.1 \\
 & CLIP-Adapter \cite{gao2024clip} & 64.2 & 69.3 & 23.5 & 55.2 & 65.4 & 81.5 & 89.3 & 78.0 & 93.9 & 50.8 & 73.0 & 67.6 \\
 & KgCoOp \cite{yao2023visual} & 65.1 & 69.5 & 24.7 & 66.2 & 65.0 & {81.7} & 90.3 & 83.1 & 94.5 & 61.1 & 74.7 & 70.5 \\
 & TaskRes \cite{yu2023task} & {67.4} & 71.9 & 31.9 & 74.9 & 73.8 & 80.6 & 89.1 & {93.5} & {94.8} & {64.5} & 78.4 & 74.6 \\
 & MaPLe \cite{khattak2023maple} & 66.3 & 70.3 & 25.4 & 79.0 & 63.7 & {81.9} & {90.9} & 81.1 & 94.4 & 59.8 & 75.0 & 71.6 \\
 & ProGrad \cite{zhu2023prompt} & 66.1 & 71.1 & 29.0 & 73.5 & 71.8 & 80.0 & 89.1 & 92.1 & 94.2 & 62.3 & 75.7 & 73.2 \\
 & LP++ \cite{huang2024lp++} & 67.1 & {72.2} & 30.3 & 78.8 & 71.2 & 81.5 & 89.3 & 92.4 & 94.6 & 64.2 & 78.4 & 74.5 \\
 & CLIP-LoRA \cite{zanella2024low} & {67.2} & 72.1 & {36.1} & {88.8} & {74.4} & 76.7 & 87.7 & 92.4 & {94.8} & 63.7 & {80.1} & {75.8} \\
 & MMA \cite{yang2024mma} & 66.7 & {72.2} & 29.6 & 56.2 & 70.4 & 81.0 & {91.0} & 90.7 & 94.6 & 64.4 & 78.7 & 72.3 \\
\rowcolor{lightgray!30} \cellcolor{white} &
 MMLoP & 66.2 & 72.2 & 31.4 & 85.2 & 71.0 & 81.2 & 90.8 & 92.6 & 95.0 & 65.7 & 79.7 & {75.5}\\ \\ [-5ex] \\ 
    \cmidrule(lr){1-14}
    \cellcolor{gray!0} \multirow{13}{*}{4} & \emph{Zero-Shot}  & 61.9 & 62.0 & 19.3 & 45.1 & 60.4 & 80.5 & 87.5 & 67.0 & 91.1 & 42.6 & 62.2 & 61.8 \\
 & CoOp \cite{COOP} (ctx=16) & 63.2 & 67.1 & 24.0 & 68.7 & 66.2 & 75.6 & 88.8 & 87.9 & 93.0 & 55.3 & 75.0 & 69.5 \\
 & CoCoOp \cite{cocoop} & 65.2 & 67.8 & 17.3 & 58.5 & 62.0 & 81.1 & 89.8 & 74.6 & 93.2 & 52.3 & 71.6 & 66.7 \\
 & TIP-Adapter-F \cite{zhang2021tip} & 65.8 & 68.3 & {28.8} & 71.5 & 67.6 & 80.9 & 88.6 & 88.9 & {94.6} & 58.0 & 75.1 & 71.6 \\
 & CLIP-Adapter \cite{gao2024clip} & 63.7 & 65.6 & 21.3 & 49.9 & 62.2 & {81.3} & 88.4 & 68.3 & 92.0 & 47.2 & 67.3 & 64.3 \\
 & KgCoOp \cite{yao2023visual} & 64.7 & 69.2 & 22.6 & 64.9 & 63.2 & 81.2 & 89.5 & 76.8 & 93.8 & 55.1 & 71.6 & 68.4 \\
 & TaskRes \cite{yu2023task} & {66.1} & 66.7 & 23.1 & 70.7 & 66.7 & 76.7 & 86.7 & 79.0 & 90.6 & 57.0 & 68.2 & 68.3 \\
 & MaPLe \cite{khattak2023maple} & 65.6 & 69.4 & 23.4 & 64.7 & 62.2 & {81.4} & {90.5} & 78.1 & 94.0 & 55.0 & 70.9 & 68.7 \\
 & ProGrad \cite{zhu2023prompt} & 65.2 & 69.6 & 24.8 & 63.7 & 66.4 & 79.2 & 89.4 & 87.5 & 93.2 & 55.9 & 73.4 & 69.8 \\
 & LP++ \cite{huang2024lp++} & {66.1} & {70.5} & 26.0 & 73.5 & 67.3 & 80.0 & 88.9 & {90.2} & 94.0 & 59.3 & 74.8 & 71.9 \\
 & CLIP-LoRA \cite{zanella2024low} & {66.5} & 70.3 & 27.7 & {85.6} & 68.3 & 75.6 & 86.3 & 90.1 & 94.3 & {60.3} & {76.5} & {72.9} \\
 & MMA \cite{yang2024mma} & 64.7 & 70.4 & 25.6 & 36.0 & 66.3 & 80.5 & {90.7} & 86.1 & 94.0 & 55.6 & 74.6 & 67.7 \\
\rowcolor{lightgray!30} \cellcolor{white} &
 MMLoP & 65.6 & 70.5 & 27.6 & 80.7 & 68.2 & 79.8 & 90.3 & 89.8 & 94.5 & 60.0 & 76.3 & {73.0}\\ \\ [-5ex] \\ 
 \cmidrule(lr){1-14}
    \end{tabular}
    \end{adjustbox}
\label{tab:vit32}
\end{table*}

\clearpage
\begin{table}[tp]
\centering
\setlength{\abovecaptionskip}{0.15cm}  
\caption{Per-dataset base-to-novel generalization comparison with low-rank baselines. HM refers to harmonic mean.}
\label{tab:base_to_novel_lowrank}
\renewcommand\arraystretch{1.1}
\setlength{\tabcolsep}{1.5mm}{
\resizebox{\textwidth}{!}{
    \begin{tabular}{|l|ccc|ccc|ccc|ccc|}
    \hline
    \multirow{2}{*}{\textbf{Method}} &
      \multicolumn{3}{c|}{\textbf{Average}} &
      \multicolumn{3}{c|}{\textbf{ImageNet}} &
      \multicolumn{3}{c|}{\textbf{Caltech101}} &
      \multicolumn{3}{c|}{\textbf{OxfordPets}} \\ \cline{2-13}
     & Base & Novel & HM & Base & Novel & HM & Base & Novel & HM & Base & Novel & HM \\ \hline
    DIP~\cite{hao2023towards}            & 79.86 & 74.17 & 76.91 & 76.07 & 70.87 & 73.37 & 97.90 & 95.03 & 96.45 & 95.17 & 97.77 & 96.45 \\
    CLIP-LoRA~\cite{zanella2024low}      & 85.32 & 70.63 & 77.28 & 77.58 & 68.76 & 72.91 & 98.19 & 93.05 & 95.55 & 94.36 & 95.71 & 95.03 \\
    Block-LoRA~\cite{zhou2025one}     & 85.22 & 73.32 & 78.82 & 76.17 & 65.40 & 70.37 & 98.33 & 94.27 & 96.26 & 95.33 & 97.70 & 96.50 \\
    Comp-LoRA~\cite{wang2025complementary}      & 85.24 & 73.22 & 78.77 & 76.00 & 65.63 & 70.44 & 98.30 & 93.87 & 96.03 & 95.40 & 97.70 & 96.54 \\
    \cline{1-13}
    \rowcolor{lightgray!30}
    MMLoP (Ours)        & 83.79 & 75.98 & 79.70 & 77.00 & 70.50 & 73.61 & 98.27 & 93.93 & 96.05 & 95.47 & 96.97 & 96.21 \\ \hline
    \hline
    \multirow{2}{*}{\textbf{Method}} &
      \multicolumn{3}{c|}{\textbf{StanfordCars}} &
      \multicolumn{3}{c|}{\textbf{Flowers102}} &
      \multicolumn{3}{c|}{\textbf{Food101}} &
      \multicolumn{3}{c|}{\textbf{FGVCAircraft}} \\ \cline{2-13}
     & Base & Novel & HM & Base & Novel & HM & Base & Novel & HM & Base & Novel & HM \\ \hline
    DIP~\cite{hao2023towards}            & 70.00 & 75.10 & 72.46 & 94.80 & 74.97 & 83.72 & 90.73 & 91.83 & 91.28 & 35.17 & 35.57 & 35.37 \\
    CLIP-LoRA~\cite{zanella2024low}      & 83.93 & 65.54 & 73.60 & 97.91 & 68.61 & 80.68 & 86.84 & 86.67 & 86.76 & 50.10 & 26.03 & 34.26 \\
    Block-LoRA~\cite{zhou2025one}     & 83.03 & 69.03 & 75.39 & 97.97 & 73.07 & 83.70 & 89.53 & 90.47 & 90.00 & 49.20 & 31.13 & 38.14 \\
    Comp-LoRA~\cite{wang2025complementary}      & 82.63 & 68.17 & 74.71 & 97.97 & 72.50 & 83.33 & 89.17 & 90.37 & 89.76 & 49.13 & 30.43 & 37.59 \\
    \cline{1-13}
    \rowcolor{lightgray!30}
    MMLoP (Ours)        & 77.77 & 74.83 & 76.27 & 97.63 & 76.73 & 85.93 & 90.70 & 91.70 & 91.19 & 42.17 & 34.60 & 38.01 \\ \hline
    \hline
    \multirow{2}{*}{\textbf{Method}} &
      \multicolumn{3}{c|}{\textbf{SUN397}} &
      \multicolumn{3}{c|}{\textbf{DTD}} &
      \multicolumn{3}{c|}{\textbf{EuroSAT}} &
      \multicolumn{3}{c|}{\textbf{UCF101}} \\ \cline{2-13}
     & Base & Novel & HM & Base & Novel & HM & Base & Novel & HM & Base & Novel & HM \\ \hline
    DIP~\cite{hao2023towards}            & 79.47 & 77.70 & 78.57 & 74.27 & 60.33 & 66.58 & 81.57 & 61.67 & 70.23 & 83.33 & 75.03 & 78.97 \\
    CLIP-LoRA~\cite{zanella2024low}      & 81.11 & 74.53 & 77.68 & 83.95 & 62.84 & 71.39 & 97.04 & 62.50 & 76.03 & 87.52 & 72.74 & 79.45 \\
    Block-LoRA~\cite{zhou2025one}     & 81.57 & 76.13 & 78.76 & 83.70 & 64.00 & 72.54 & 94.87 & 67.17 & 78.65 & 87.73 & 78.10 & 82.64 \\
    Comp-LoRA~\cite{wang2025complementary}      & 81.43 & 75.63 & 78.43 & 84.00 & 64.37 & 72.88 & 95.47 & 69.00 & 80.10 & 88.13 & 77.77 & 82.63 \\
    \cline{1-13}
    \rowcolor{lightgray!30}
    MMLoP (Ours)        & 82.40 & 78.13 & 80.21 & 82.67 & 60.87 & 70.11 & 92.37 & 79.10 & 85.22 & 85.23 & 78.37 & 81.66 \\ \hline
    \end{tabular}
}
}
\end{table}
\section{Comparison with Low-Rank Adaptation Methods}
\label{low-rank-methods}

We compare MMLoP against four representative low-rank baselines on base-to-novel
generalization (Tab.~\ref{tab:base_to_novel_lowrank}). CLIP-LoRA~\cite{zanella2024low},
Block-LoRA~\cite{zhou2025one}, and Comp-LoRA~\cite{wang2025complementary} are weight-space methods that
inject low-rank adapters into the frozen CLIP backbone, while DIP~\cite{hao2023towards} is a
prompt-space method applying low-rank constraints to shallow text-side prompts.
\textbf{MMLoP achieves the highest HM among all four}, with substantially
stronger novel-class accuracy ($+2.66$ over Block-LoRA, $+2.76$ over Comp-LoRA,
$+5.35$ over CLIP-LoRA, and $+1.81$ over DIP). While the three weight-space
baselines reach higher base accuracy ($\sim$$85.2$), they overfit to base classes
and lose generalization---precisely the phenomenon our SCL and UDC components are
designed to prevent.

\section{Per-Dataset Ablation Study}
\subsection{Ablation on Consistency Loss Formulation}

Table~\ref{tab:scl_ablation} compares symmetric ($\mathcal{L}_{SKL}$) and 
asymmetric ($\mathcal{L}_{KL}$) formulations of the logit-level 
consistency loss, both with and without Uniform Drift Correction. 
The symmetric formulation consistently achieves higher novel accuracy 
and harmonic mean across datasets, confirming our design choice in 
Eq.~(6). UDC provides complementary gains under both loss variants.

\begin{table*}[h]
\caption{Comparison of symmetric ($\mathcal{L}_{\text{SKL}}$) and asymmetric 
($\mathcal{L}_{\text{KL}}$) consistency losses, with and without Uniform Drift 
Correction (UDC), averaged over 9 datasets. HM refers to harmonic mean.}
\label{tab:scl_ablation}
\small \centering
\renewcommand\arraystretch{1.2}
\setlength{\tabcolsep}{5pt}
\scalebox{0.7}[0.7]{
\begin{tabular}{|l|cc|c|cc|c|cc|c|cc|c|}
\hline
\multirow{2}{*}{\textbf{Dataset}} & \multicolumn{3}{c|}{\textbf{$\mathcal{L}_\text{SKL}$}} & \multicolumn{3}{c|}{\textbf{$\mathcal{L}_\text{SKL} + \text{UDC}$}} & \multicolumn{3}{c|}{\textbf{$\mathcal{L}_\text{KL}$}} & \multicolumn{3}{c|}{\textbf{$\mathcal{L}_\text{KL} + \text{UDC}$}} \\ \cline{2-13}
 & \textbf{Base} & \textbf{Novel} & \textbf{HM} & \textbf{Base} & \textbf{Novel} & \textbf{HM} & \textbf{Base} & \textbf{Novel} & \textbf{HM} & \textbf{Base} & \textbf{Novel} & \textbf{HM} \\ \hline
Caltech101    & 98.10 & 94.07 & 96.04 & 98.27 & 93.93 & 96.05 & 98.20 & 93.80 & 95.95 & 98.17 & 94.57 & 96.34 \\
DTD           & 82.47 & 61.97 & 70.77 & 82.67 & 60.87 & 70.11 & 82.20 & 61.27 & 70.21 & 82.57 & 60.23 & 69.65 \\
EuroSAT       & 90.90 & 75.37 & 82.41 & 92.37 & 79.10 & 85.23 & 92.37 & 72.50 & 81.24 & 91.70 & 77.23 & 83.85 \\
FGVCAircraft  & 41.30 & 34.70 & 37.71 & 42.17 & 34.60 & 38.00 & 41.30 & 34.47 & 37.58 & 41.40 & 34.63 & 37.71 \\
Food101       & 90.77 & 91.47 & 91.12 & 90.70 & 91.70 & 91.20 & 90.73 & 91.50 & 91.11 & 90.73 & 91.67 & 91.20 \\
OxfordFlowers & 97.50 & 76.17 & 85.53 & 97.63 & 76.73 & 85.93 & 97.50 & 76.70 & 85.86 & 97.53 & 75.13 & 84.88 \\
OxfordPets    & 95.40 & 97.27 & 96.33 & 95.47 & 96.97 & 96.21 & 95.43 & 97.37 & 96.39 & 95.37 & 97.27 & 96.31 \\
StanfordCars  & 77.27 & 75.07 & 76.15 & 77.77 & 74.83 & 76.27 & 77.43 & 75.13 & 76.26 & 77.63 & 74.73 & 76.15 \\
UCF101        & 86.23 & 76.60 & 81.13 & 85.23 & 78.37 & 81.66 & 85.57 & 77.93 & 81.57 & 85.97 & 78.60 & 82.12 \\ \hline
\textbf{Average} & 84.44 & 75.85 & 79.91 & 84.70 & 76.34 & 80.30 & 84.53 & 75.63 & 79.83 & 84.56 & 76.01 & 80.06 \\ \hline
\end{tabular}
}
\end{table*}

\subsection{Per-Dataset Breakdown of Incremental Ablation Study}

Table~\ref{tab_appendix:base_to_new} reports the per-dataset breakdown of 
the incremental ablation study summarized in Table~4 of the main body.  The results confirm that the trends observed on average hold consistently across the majority of datasets. The Self-Regulating Consistency Loss ($\mathcal{L}_{\text{SCL}}$) provides the largest gains in novel class 
accuracy across most datasets. Uniform Drift Correction (UDC) further improves novel 
accuracy across nearly all datasets, with the largest gains on EuroSAT 
and UCF101. The Shared Up-Projection consistently improves or maintains 
the harmonic mean, confirming that cross-modal alignment provides complementary benefits to feature-level regularization across diverse recognition tasks.

\begin{table*}[!t]
\caption{Detailed ablation study on base and novel class accuracy across 11 
datasets, showing the incremental contribution of each proposed component. 
$\Delta$ denotes the change in harmonic mean relative to the IVLP baseline. 
HM refers to harmonic mean.}
\label{tab_appendix:base_to_new}
\small\centering
\renewcommand\arraystretch{1.1}
\setlength{\tabcolsep}{10pt}
\resizebox{\textwidth}{!}{
\begin{tabular}{|lc|ccccc|c|}
\hline
\textbf{Dataset} & & \textbf{IVLP} & \textbf{+ LR Prompting} & \textbf{+ $\mathcal{L}_\text{SCL}$} & \textbf{+ UDC} & \textbf{+ Shared Up-Proj.} & \textbf{$\Delta$} \\ \hline
\multirow{3}{*}{Average over 11 datasets}
  & Base Acc.  & 84.21 & 83.70 & 83.60 & 83.78 & 83.79 & \textcolor{red}{-0.42} \\
  & Novel Acc. & 71.79 & 71.39 & 74.48 & 75.12 & 75.98 & \textcolor{teal}{+4.19} \\
  & H.M        & 77.51 & 77.06 & 78.78 & 79.20 & 79.70 & \textcolor{teal}{+2.19} \\ \hline
\multirow{3}{*}{ImageNet}
  & Base Acc.  & 77.00 & 77.00 & 77.10 & 77.07 & 77.00 & \textcolor{teal}{+0.00} \\
  & Novel Acc. & 66.50 & 67.30 & 70.27 & 70.67 & 70.50 & \textcolor{teal}{+4.00} \\
  & H.M        & 71.37 & 71.82 & 73.53 & 73.76 & 73.61 & \textcolor{teal}{+2.24} \\ \hline
\multirow{3}{*}{Caltech101}
  & Base Acc.  & 98.30 & 98.23 & 98.17 & 98.20 & 98.27 & \textcolor{teal}{-0.03} \\
  & Novel Acc. & 93.20 & 93.63 & 94.03 & 94.20 & 93.93 & \textcolor{teal}{+0.73} \\
  & H.M        & 95.68 & 95.87 & 96.06 & 96.15 & 96.05 & \textcolor{teal}{+0.37} \\ \hline
\multirow{3}{*}{OxfordPets}
  & Base Acc.  & 94.90 & 94.97 & 95.30 & 95.50 & 95.47 & \textcolor{teal}{+0.57} \\
  & Novel Acc. & 97.20 & 97.10 & 97.23 & 97.37 & 96.97 & \textcolor{teal}{-0.23} \\
  & H.M        & 96.04 & 96.02 & 96.26 & 96.43 & 96.21 & \textcolor{teal}{+0.17} \\ \hline
\multirow{3}{*}{StanfordCars}
  & Base Acc.  & 79.53 & 78.90 & 77.40 & 77.70 & 77.77 & \textcolor{red}{-1.76} \\
  & Novel Acc. & 71.47 & 70.33 & 74.73 & 74.27 & 74.83 & \textcolor{teal}{+3.36} \\
  & H.M        & 75.28 & 74.37 & 76.04 & 75.97 & 76.27 & \textcolor{teal}{+0.99} \\ \hline
\multirow{3}{*}{Flowers102}
  & Base Acc.  & 97.97 & 97.83 & 97.77 & 97.97 & 97.63 & \textcolor{red}{-0.34} \\
  & Novel Acc. & 72.10 & 70.67 & 74.47 & 75.73 & 76.73 & \textcolor{teal}{+4.63} \\
  & H.M        & 83.07 & 82.06 & 84.54 & 85.39 & 85.93 & \textcolor{teal}{+2.86} \\ \hline
\multirow{3}{*}{Food101}
  & Base Acc.  & 89.37 & 89.40 & 90.83 & 90.93 & 90.70 & \textcolor{teal}{+1.33} \\
  & Novel Acc. & 90.30 & 89.20 & 91.57 & 91.77 & 91.70 & \textcolor{teal}{+1.40} \\
  & H.M        & 89.83 & 89.30 & 91.20 & 91.35 & 91.19 & \textcolor{teal}{+1.36} \\ \hline
\multirow{3}{*}{FGVCAircraft}
  & Base Acc.  & 42.60 & 43.30 & 40.47 & 41.03 & 42.17 & \textcolor{red}{-0.43} \\
  & Novel Acc. & 25.23 & 31.73 & 34.07 & 34.07 & 34.60 & \textcolor{teal}{+9.37} \\
  & H.M        & 31.69 & 36.62 & 37.00 & 37.24 & 38.01 & \textcolor{teal}{+6.32} \\ \hline
\multirow{3}{*}{SUN397}
  & Base Acc.  & 81.60 & 81.57 & 82.10 & 82.07 & 82.40 & \textcolor{teal}{+0.80} \\
  & Novel Acc. & 75.50 & 74.17 & 77.30 & 78.53 & 78.13 & \textcolor{teal}{+2.63} \\
  & H.M        & 78.43 & 77.69 & 79.63 & 80.29 & 80.21 & \textcolor{teal}{+1.78} \\ \hline
\multirow{3}{*}{DTD}
  & Base Acc.  & 82.40 & 80.83 & 83.30 & 82.60 & 82.67 & \textcolor{teal}{+0.27} \\
  & Novel Acc. & 56.20 & 53.80 & 57.33 & 58.60 & 60.87 & \textcolor{teal}{+4.67} \\
  & H.M        & 66.82 & 64.60 & 67.92 & 68.56 & 70.11 & \textcolor{teal}{+3.29} \\ \hline
\multirow{3}{*}{EuroSAT}
  & Base Acc.  & 96.73 & 94.03 & 91.20 & 92.43 & 92.37 & \textcolor{red}{-4.36} \\
  & Novel Acc. & 67.83 & 64.40 & 69.97 & 71.83 & 79.10 & \textcolor{teal}{+11.27} \\
  & H.M        & 79.74 & 76.44 & 79.19 & 80.84 & 85.22 & \textcolor{teal}{+5.48} \\ \hline
\multirow{3}{*}{UCF101}
  & Base Acc.  & 85.93 & 84.63 & 85.90 & 86.03 & 85.23 & \textcolor{red}{-0.70} \\
  & Novel Acc. & 74.17 & 72.93 & 78.27 & 79.27 & 78.37 & \textcolor{teal}{+4.20} \\
  & H.M        & 79.62 & 78.35 & 81.91 & 82.50 & 81.66 & \textcolor{teal}{+2.04} \\ \hline
\end{tabular}
}
\end{table*}

\subsection{Per-Dataset Hyperparameter Sensitivity Analysis}

Tables~\ref{tab:lora_rank}, \ref{tab:prompt_length}, and~\ref{tab:prompt_depth} report 
the per-dataset breakdown of the hyperparameter sensitivity analysis 
summarized in Table~5 of the main paper, covering prompt factorization rank, prompt 
length, and prompt depth respectively. Note that SUN397 and ImageNet 
are excluded from these tables due to their computational cost; however, 
based on our experiments we found that both datasets exhibit low 
sensitivity to these hyperparameters, consistent with the trends 
observed across the remaining datasets. The results confirm that the 
trends observed on average are consistent across individual datasets. 
For prompt factorization rank, rank-1 achieves the best novel accuracy across the 
majority of datasets, with higher ranks occasionally improving base 
accuracy at the cost of generalization to novel classes. For prompt 
length, performance is relatively stable around the chosen value of 4, 
with no single dataset showing large sensitivity to this parameter. 
For prompt depth, depth 9 provides the best balance between base and 
novel accuracy across most datasets, while depth 11 tends to improve 
base accuracy but hurts novel generalization, consistent with the 
overfitting behavior discussed in Section~5.4. Overall, these results confirm that MMLoP is robust to hyperparameter choices across diverse 
datasets and recognition tasks, and that all hyperparameters can be fixed globally without dataset-specific tuning.

\begin{table*}[!t]
\caption{Effect of LoRA rank on base and novel class accuracy. Results are averaged over 3 seeds. HM refers to harmonic mean.}
\label{tab:lora_rank}
\small \centering
\renewcommand\arraystretch{1.2}
\setlength{\tabcolsep}{6pt}
\scalebox{0.8}[0.8]{
\begin{tabular}{|l|cc|c|cc|c|cc|c|}
\hline
 & \multicolumn{3}{c|}{\textbf{Rank 1}} & \multicolumn{3}{c|}{\textbf{Rank 2}} & \multicolumn{3}{c|}{\textbf{Rank 4}} \\ \cline{2-10}
\textbf{Dataset} & \textbf{Base} & \textbf{Novel} & \textbf{HM} & \textbf{Base} & \textbf{Novel} & \textbf{HM} & \textbf{Base} & \textbf{Novel} & \textbf{HM} \\ \hline
Caltech101 & 98.10 & 94.07 & 96.04 & 98.07 & 93.80 & 95.89 & 98.07 & 94.17 & 96.08 \\ 
DTD & 82.47 & 61.97 & 70.77 & 82.60 & 56.33 & 66.98 & 81.80 & 59.87 & 69.13 \\
EuroSAT & 90.90 & 75.37 & 82.41 & 91.23 & 72.27 & 80.66 & 91.47 & 70.77 & 79.82 \\ 
FGVCAircraft & 41.30 & 34.70 & 37.71 & 42.00 & 35.67 & 38.58 & 41.93 & 34.97 & 38.15 \\ 
Food101 & 90.77 & 91.47 & 91.12 & 90.77 & 91.57 & 91.17 & 90.80 & 91.67 & 91.23 \\ 
OxfordFlowers & 97.50 & 76.17 & 85.53 & 97.83 & 76.27 & 85.71 & 98.13 & 76.47 & 85.93 \\ 
OxfordPets & 95.40 & 97.27 & 96.33 & 95.27 & 96.73 & 95.99 & 95.17 & 97.13 & 96.14 \\ 
StanfordCars & 77.27 & 75.07 & 76.15 & 78.37 & 74.30 & 76.28 & 79.10 & 74.03 & 76.48 \\ 
UCF101 & 86.23 & 76.60 & 81.13 & 86.73 & 78.17 & 82.23 & 86.73 & 77.73 & 81.98 \\ \hline
\textbf{Average} & 84.44 & 75.85 & 79.91 & 84.76 & 75.01 & 79.59 & 84.80 & 75.20 & 79.71 \\ \hline
\end{tabular}
}
\end{table*}

\begin{table*}[t]
\caption{Effect of prompt length on base and novel class accuracy. Results are averaged over 3 seeds. HM refers to harmonic mean.}
\label{tab:prompt_length}
\small \centering
\renewcommand\arraystretch{1.2}
\setlength{\tabcolsep}{6pt}
\scalebox{0.8}[0.8]{
\begin{tabular}{|l|cc|c|cc|c|cc|c|}
\hline
 & \multicolumn{3}{c|}{\textbf{Length 2}} & \multicolumn{3}{c|}{\textbf{Length 4}} & \multicolumn{3}{c|}{\textbf{Length 8}} \\ \cline{2-10}
\textbf{Dataset} & \textbf{Base} & \textbf{Novel} & \textbf{HM} & \textbf{Base} & \textbf{Novel} & \textbf{HM} & \textbf{Base} & \textbf{Novel} & \textbf{HM} \\ \hline
Caltech101 & 98.00 & 94.37 & 96.15 & 98.10 & 94.07 & 96.04 & 98.27 & 93.43 & 95.79 \\ 
DTD & 82.23 & 60.60 & 69.78 & 82.47 & 61.97 & 70.77 & 82.77 & 58.10 & 68.27 \\
EuroSAT & 91.20 & 66.57 & 76.96 & 90.90 & 75.37 & 82.41 & 92.23 & 66.30 & 77.14 \\ 
FGVCAircraft & 39.87 & 34.60 & 37.05 & 41.30 & 34.70 & 37.71 & 42.33 & 34.83 & 38.22 \\ 
Food101 & 90.67 & 91.53 & 91.10 & 90.77 & 91.47 & 91.12 & 90.73 & 91.53 & 91.13 \\ 
OxfordFlowers & 97.40 & 76.90 & 85.94 & 97.50 & 76.17 & 85.53 & 97.60 & 76.10 & 85.52 \\ 
OxfordPets & 95.30 & 97.20 & 96.24 & 95.40 & 97.27 & 96.33 & 95.43 & 97.00 & 96.21 \\ 
StanfordCars & 76.17 & 74.97 & 75.57 & 77.27 & 75.07 & 76.15 & 77.40 & 74.40 & 75.87 \\ 
UCF101 & 86.60 & 78.17 & 82.17 & 86.23 & 76.60 & 81.13 & 86.67 & 77.70 & 81.94 \\ \hline
\textbf{Average} & 84.16 & 74.99 & 79.31 & 84.44 & 75.85 & 79.91 & 84.83 & 74.38 & 79.26 \\ \hline
\end{tabular}
}
\end{table*}

\begin{table*}[!t]
\caption{Effect of prompt depth on base and novel class accuracy. Results are averaged over 3 seeds. HM refers to harmonic mean.}
\label{tab:prompt_depth}
\small \centering
\renewcommand\arraystretch{1.2}
\setlength{\tabcolsep}{6pt}
\scalebox{0.8}[0.8]{
\begin{tabular}{|l|cc|c|cc|c|cc|c|}
\hline
 & \multicolumn{3}{c|}{\textbf{Depth 7}} & \multicolumn{3}{c|}{\textbf{Depth 9}} & \multicolumn{3}{c|}{\textbf{Depth 11}} \\ \cline{2-10}
\textbf{Dataset} & \textbf{Base} & \textbf{Novel} & \textbf{HM} & \textbf{Base} & \textbf{Novel} & \textbf{HM} & \textbf{Base} & \textbf{Novel} & \textbf{HM} \\ \hline
Caltech101 & 98.00 & 94.00 & 95.96 & 98.10 & 94.07 & 96.04 & 98.10 & 93.93 & 95.97 \\ 
DTD & 81.10 & 59.63 & 68.73 & 82.47 & 61.97 & 70.77 & 83.27 & 59.53 & 69.43 \\
EuroSAT & 90.27 & 65.53 & 75.94 & 90.90 & 75.37 & 82.41 & 91.17 & 67.37 & 77.48 \\ 
FGVCAircraft & 39.47 & 33.03 & 35.96 & 41.30 & 34.70 & 37.71 & 42.23 & 34.90 & 38.22 \\ 
Food101 & 90.77 & 91.50 & 91.13 & 90.77 & 91.47 & 91.12 & 90.63 & 91.67 & 91.15 \\ 
OxfordFlowers & 97.43 & 75.87 & 85.31 & 97.50 & 76.17 & 85.53 & 97.93 & 75.77 & 85.44 \\ 
OxfordPets & 95.03 & 96.60 & 95.81 & 95.40 & 97.27 & 96.33 & 95.23 & 97.07 & 96.14 \\ 
StanfordCars & 75.27 & 74.67 & 74.97 & 77.27 & 75.07 & 76.15 & 78.00 & 73.53 & 75.70 \\ 
UCF101 & 85.30 & 76.87 & 80.87 & 86.23 & 76.60 & 81.13 & 87.27 & 77.43 & 82.06 \\ \hline
\textbf{Average} & 83.63 & 74.19 & 78.63 & 84.44 & 75.85 & 79.91 & 84.87 & 74.58 & 79.39 \\ \hline
\end{tabular}
}
\end{table*}

\end{document}